\documentclass[a4paper,fleqn,preprint]{cas-sc}

\usepackage[authoryear,longnamesfirst]{natbib}
\usepackage{tabularx}
\usepackage{textcomp, gensymb}
\usepackage{rotating}
\usepackage{lineno}
\usepackage{placeins}
\usepackage{graphicx}
\usepackage{nomencl}
\makenomenclature

\def\tsc#1{\csdef{#1}{\textsc{\lowercase{#1}}\xspace}}
\tsc{WGM}
\tsc{QE}
\tsc{EP}
\tsc{PMS}
\tsc{BEC}
\tsc{DE}

\begin{document}
\let\WriteBookmarks\relax
\def\floatpagepagefraction{1}
\def\textpagefraction{.001}
\def\code#1{\texttt{#1}}

\shorttitle{FORWARD: Dataset of a forwarder operating in rough terrain}

\shortauthors{Lundbäck et~al.}

\title [mode = title]{FORWARD: Dataset of a forwarder operating in rough terrain}

\author[1]{Mikael Lundbäck}[orcid=0000-0002-1842-7032]
\author[1]{Erik Wallin}[orcid=0000-0001-6266-4740]
\author[2]{Carola Häggström}[orcid=0000-0002-6351-4469]
\author[3]{Mattias Nyström}[orcid=0000-0002-5872-6862]
\author[3]{Andreas Grönlund}[orcid=0000-0002-1543-7358]
\author[4]{Mats Richardson}[orcid=0009-0005-0982-1897]
\author[4]{Petrus Jönsson}
\author[2]{William Arnvik}[orcid=0009-0008-9673-2236]
\author[1]{Lucas Hedström}[orcid=0000-0002-3315-0633]
\author[1]{Arvid Fälldin}[orcid=0009-0000-9267-1140]
\author[1]{Martin Servin*}[orcid=0000-0002-0787-4988]
\ead{martin.servin@umu.se}

\credit{Conceptualization of this study, Methodology, Software}

\affiliation[1]{organization={Umeå University},
    addressline={Department of Physics},
    city={Umeå},
    postcode={SE-90187},
    country={Sweden}}

\affiliation[2]{organization={Swedish University of Agricultural Sciences},
    addressline={Department of Forest Biomaterials and Technology},
    city={Umeå},
    postcode={SE-90183},
    country={Sweden}}

\affiliation[3]{organization={Komatsu Forest AB},
    city={Umeå},
    postcode={SE-90137},
    country={Sweden}}

\affiliation[3]{organization={Skogforsk, Swedish Forest Research Institute},
    city={Uppsala},
    postcode={SE-75183},
    country={Sweden}}

\nonumnote{This work was supported in part by Mistra Digital Forest (Grant DIA 2017/14 \#6) and Horizon Europe Project XSCAVE under Grant 101189836.}

\begin{abstract}
    We present FORWARD, a high-resolution multimodal dataset of a cut-to-length forwarder operating in rough terrain on two harvest sites in the middle part of Sweden. The forwarder is a large Komatsu model equipped with vehicle telematics sensors, including global positioning via satellite navigation, movement sensors, accelerometers, and engine sensors. The forwarder was additionally equipped with cameras, operator vibration sensors, and multiple Inertial Measurement Units (IMUs). The data includes event time logs recorded at 5 Hz of driving speed, fuel consumption, machine position with centimeter accuracy, and crane use while the forwarder operates in forest areas, aerially laser-scanned with a resolution of around 1500 points per square meter. Production log files (Standard for Forestry Data, StanForD) with time-stamped machine events, extensive video material, and terrain data in various formats are included as well. About 18 hours of regular wood extraction work during three days is annotated from 360\textdegree-video material into individual work elements and included in the dataset. We also include scenario specifications of conducted experiments on forest roads and in terrain. Scenarios include repeatedly driving the same routes with and without steel tracks, different load weights, and different target driving speeds. The dataset is intended for developing models and algorithms for trafficability, perception, and autonomous control of forest machines using artificial intelligence, simulation, and experiments on physical testbeds. In part, we focus on forwarders traversing terrain, avoiding or handling obstacles, and loading or unloading logs, with consideration for efficiency, fuel consumption, safety, and environmental impact. Other benefits of the open dataset include the ability to explore auto-generation and calibration of forestry machine simulators and automation scenario descriptions using the data recorded in the field. The data and scripts for data exploration and analysis are made long-term publicly available through the Swedish National Data Service.
\end{abstract}



\ExplSyntaxOn
\keys_set:nn { stm / mktitle } { nologo }
\ExplSyntaxOff
\maketitle

\newpage
\nomenclature{\(IMU\)}{Inertial Measurement Unit}
\nomenclature{\(StanForD\)}{Standard for Forestry Data}
\nomenclature{\(RTK-GNSS\)}{Real Time Kinematic Global Navigation Satellite System}
\nomenclature{\(CAN\)}{Controller Area Network}
\nomenclature{\(LiDAR\)}{Light Detection and Ranging}

\printnomenclature

\newpage
\section*{Specification Table}
\begin{table}[width=1.0\linewidth,cols=4,pos=h!]
\begin{tabularx}{\textwidth}{lX}
\toprule
Subject & Engineering \& Materials science \\
\midrule
Specific subject area & Forest Operations, Automation of Heavy Mobile Equipment, Field Robotics, Machine Learning,  Physics-Based Simulation. \\
\midrule
Type of data & In selection: Vehicle telematics data as time-series (.csv), terrain data (.las, .tif), photos and video data (.360, .mp4, .mov, .jpg), Inertial Measurement Unit (IMU) data (.csv), Standard for Forestry Data (StanForD) production data (.hpr, .fpr, .mom), vibration data (.wav), annotations and descriptions in table format (.csv, .xlsx), analysis code (python scripts). \\
\vspace{2mm}
& Raw, Filtered, Processed, and with scripts to analyse. \\
\midrule
Data collection & Data were collected in close connection to regular wood extraction at two harvesting sites. The machine was a large Komatsu forwarder with integrated GNSS-RTK positioning and extra sensors such as wheel-mounted IMU:s, 360\textdegree-camera, and operator vibration sensor. Sites were LiDAR-scanned (Riegl Vux 120) with helicopter prior to the harvest and photographed with drone post-harvest but pre-extraction. \\
\midrule
Data source location
 & The data were collected on the two harvesting sites Märrviken (2023) and Björsjö (2024) in Ånge municipality in the middle of Sweden. \\
\midrule
Data accessibility &
Repository name: Swedish National Data Service (SND), Researchdata.se

\vspace{2mm}
Data identification number: doi:10.71540/89rs-s553

\vspace{2mm}
Direct URL to data: https://doi.org/10.71540/89rs-s553

\vspace{2mm}
Instructions for accessing these data: Review provided README document in the data repository for instructions on download and usage of the data.
 \\
\midrule
Related research article\ \ & N/A \\
\bottomrule
\end{tabularx}
\end{table}

\section{Value of the data}
\begin{itemize}
    \item The data provides a comprehensive understanding of the dynamical behavior of a large forest machine traversing forest roads and rough terrain, passing obstacles, and loading and unloading logs. This is important for developing models and algorithms for perception, planning and autonomous control of sustainable high-precision forestry operations.
    \item With centimeter-precision positioning and multiple IMUs, the machine motion is captured in detail and can be correlated with the detailed topography of the ground. The same routes are traversed at different speeds and load weights, and with and without steel tracks on the wheels.
    \item Researchers can do traversability analysis that goes far beyond today's standard, which in forestry is limited to the local slope, surface roughness, and bearing capacity. The impact of local surface topography on fuel consumption, whole-body vibrations and effective speed can be studied. This in turn is useful for optimal route planning.
    \item The data can be used for validation and calibration of multibody dynamics simulation models which in turn are useful for motion planning and autonomous control of forest machinery.
    \item It can be analyzed how experts drive the forwarder and operate the crane in terms of slew angle in relation to how the logs are distributed over the rough terrain. From this, it is possible to extract automation scenarios and benchmarking data for testing automation solutions and extract statistics on machine utilization in relation to terrain properties.
    \item The multimodal data, including high-resolution aerial laser scanning, vehicle telematics, video, and production log files have been curated and synchronized. It enables annotation of objects in the terrain (stones, stumps, logs, vegetation). The data can thus be used for developing refined algorithms for segmenting images or 3D point clouds from forest environments.
    \item The data enables detailed analysis of what whole-body vibrations operators are exposed to under different terrain conditions.

\end{itemize}

\section{Background}
The availability of well-annotated data is essential for the development and evaluation of both data-driven and physics-based models essential for automation and high-precision planning.
The big advancements in computer vision and deep learning\textemdash including image recognition, object detection, and semantic segmentation\textemdash is largely a result of the efforts in creating and openly sharing annotated datasets.
In autonomous driving, there are several open datasets~\cite{Geiger2012CVPR,maddern20171} that support the development of algorithms for localization and mapping, and decision-making~\cite{wang2024survey}.
The same data may also be used to create realistic simulators and synthetic generation of data that complements real data and physical testing~\cite{dosovitskiy2017carla}.
In off-road robotics, terrain traversability is a key functionality for both planning and navigation.
Models for predicting the traversability of offroad vehicles and robots from local terrain features is an active field of research and there are several papers with accompanying public datasets~\cite{wigness_rugd_2019,jiang_rellis-3d_2021,triest_tartandrive_2022,sharma_cat_2022,mortimer_goose_2024,malladi2025a}.
Traversability depends highly on the size, kinematic structure, and driveline of the vehicle or robot~\cite{sharma_cat_2022}.
To the best of our knowledge there are no publicly available datasets for articulated heavy off-road vehicles such as a forest machine.
This shortage hinders the effective application of artificial intelligence and simulation in research and development towards higher levels
of automation and precision forestry, which is needed to achieve sustainability goals.
It is the aim of this work to fill this gap in open data.

Creating useful datasets is not just a matter of recording large volumes of data, e.g., logging the available data from forest machines in production.
With accurate, contextual and consistent annotation, more precise and advanced models can be created from the data, (compare~\cite{hook_data_2025, puliti_for-instance_2023}).
The data needs to have high variability.
This may include capturing data from situations that are unusual but safety-critical.
Furthermore, a useful dataset needs a well-documented structure as well as instructions and examples for the tools essential for processing the data when it is not in a standardized format.

\section{Data description}
The FORWARD dataset comprises detailed sensor and log data from a Komatsu cut-to-length forwarder, collected during operations in challenging terrain as well as on gravel roads at two Swedish harvest sites.
The complete dataset comprises about 1.1 TB and contain an extensive folder structure and array of different file types, described file-by-file in the accompanying readme-file.
Not all data modalities are available for both test sites since additional sensors were added for the second field trial (Björsjö site), what is available for respective test site can be checked in the folder structure before download.
In this section, we provide an overview of the data types and their format.

Sensor instrumentation includes machine-integrated Real Time Kinematic Global Navigation Satellite System (RTK-GNSS) for precise positioning, speed, and heading, a 360\textdegree-degree camera for visual documentation, operator seat vibration sensors, Controller Area Network (CAN-bus) signal logging, and multiple Inertial Measurement Units (IMUs) mounted on the machine.
The dataset provides time-stamped forwarder positions with centimeter-level accuracy, and machine event logs sampled at 5 Hz, covering variables such as driving speed, fuel consumption, and crane slew angle.
Additionally, high-resolution terrain data from helicopter-borne pre-harvest laser scanning (1,500 points/m$^2$), and drone footage of the harvested site is included, along with StanForD production log files containing detailed event records for each work session.

About 18 hours of regular wood extraction work during three days is annotated from 360\textdegree-video material into individual work elements.
We also include repeated experiments with altered machine configurations, both on forest roads and in terrain.
Experiments are structured and presented in a number of scenarios.
Scenarios include tests with and without steel tracks, different load weight, and different target driving speeds (inch levels) expressed as percent of maximum speed in terrain gear.
The data, scripts for data exploration and analysis, and some example analyses are publicly available through the Swedish National Data Service~\cite{Lundback2025FORWARD}.

An overview of the collected data is provided in Table~\ref{tab:overview}. More in-depth information of the data
sources and formats are described in the following sections.

\begin{table}
    \centering
    \small
    \renewcommand{\arraystretch}{1.25}
    \caption{Data overview. For the complete list of variables, see the metadata file 'Data\_variables'.}
    \begin{tabular}{|l|c|c|l|}
    \hline
    Type of data & Resolution & Size/duration & Comment \\
    \hline
    Airborne laser scan & 1,500 pt/m$^2$ & 15 ha/70 GB & Point cloud \& elevation map \\
    Drone photogrammetry  & 4,000 pt/m$^2$ & 15 ha/56 GB & Point cloud \& orthomosaic \\
    Video  & 3,840$\times$2,160 pixels at 60 fps & 373 GB & Various positions on/off machine  \\
    360\textdegree-video  & 5,376$\times$2,688 pixels at 30 fps & 20 h/560 GB & Mounted on machine roof  \\
    Manual object scanning  & n.a. & 9 objects & 3D reconstruction  \\
    GNSS position and speed & 0.1 m at 5 Hz & 110 h & By manufacturer \\
    Pose & 0.1 deg at 5 Hz & 110 h & By manufacturer  \\
    Acceleration  & 5 Hz & 110 h & By manufacturer \\
    Fuel consumption  & 5 Hz & 110 h & By manufacturer  \\
    Set speed (inch) & n.a. & 110 h & By operator \\
    Crane motion & 5 Hz & 110 h & Slew angle \\
    Ext. IMU & 5 Hz & 17 h & Mounted on wheels  \\
    Ext. vibration & 6,000 Hz & 27 h & Mounted in operator seat \\
    Production log file & n.a. & 40 h (Björsjö) & Forwarder production (fpr) \\
    Production log file & n.a. & 110 h & Harvester production (hpr) \\
    \hline
    \end{tabular}
    \label{tab:overview}
\end{table}

\subsection{Formats and file structures for time series data}
Raw data is provided in different formats depending on the data source.
Machine signals logged from the CAN-bus and GNSS receivers are provided in .csv format, with one file for each test site.
Each file contains one row per timestamp and one column per signal.
Production log files (StanForD) are provided as .fpr, .hpr, and .mom files, which are standard formats for forestry machine production data, containing time-stamped information about forwarded loads (fpr), harvested trees and processed logs (hpr), and machine positions (mom).
Each test site has its own set of production log files, separated in folders named after the test site and filetype.
The 360\textdegree-video is provided in .mp4 format as well as the original GoPro .360 format.
The .mp4-files have a resolution of 1,024$\times$592 pixels, a frame rate of 30 frames per second, and are easily played back with pan and zoom functionality in standard video players, such as VLC.
Vibration data from the operator seat is provided in .wav format, with a sampling frequency of 6,000 Hz.
There are four large files containing vibration data, spanning over the three test days at the Björsjö test site.
No high-frequency vibration data is available from the Märrviken test site.
IMU data is provided as a number of .csv files, with one file per IMU and data recording episode.

\begin{figure}[h]
    \centering
    \includegraphics[width=1.0\linewidth]{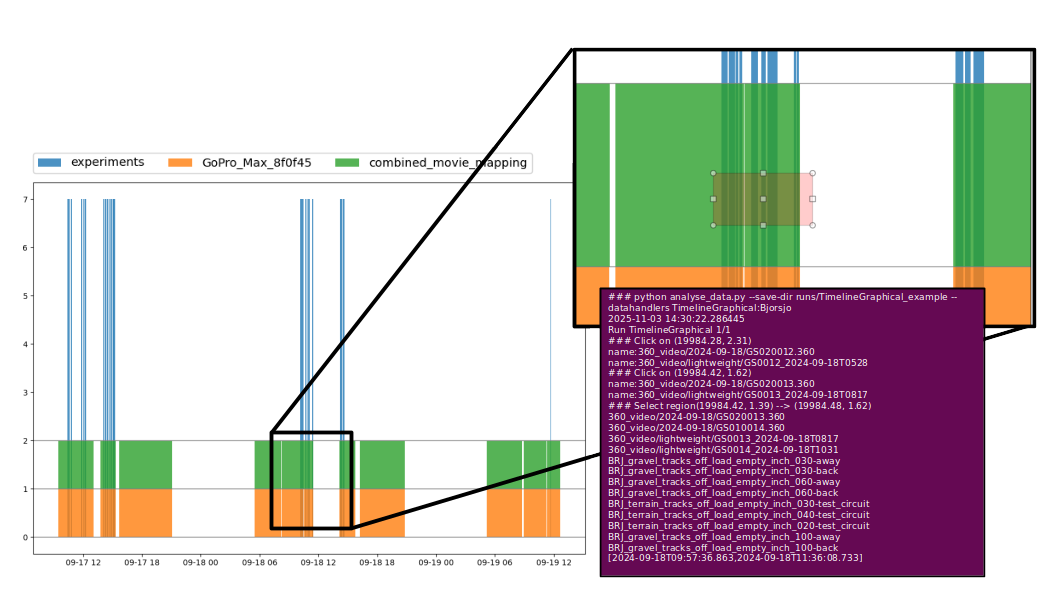}
    \caption{Interactive plot for visualizing and navigating the time series data. The user can zoom in and out, and select preferred time intervals to show experiments, video files, and exact time interval in the terminal (purple box in the figure) for further analysis. Blue lines, orange areas, and green areas indicate experiments, raw 360\textdegree-video filenames, and lightweight stitched video filenames, respectively.}
    \label{fig:interactive_plot}
\end{figure}

After loading and synchronization of the time series data, an interactive plot can be generated to visualize and navigate the data (Fig.~\ref{fig:interactive_plot}).
The practical steps to load data, synchronize time series, and generate the interactive plot are provided in the analysis scripts repository.

\subsection{Formats and file structures for terrain data}
\begin{figure}
    \centering
    \includegraphics[width=0.9\linewidth]{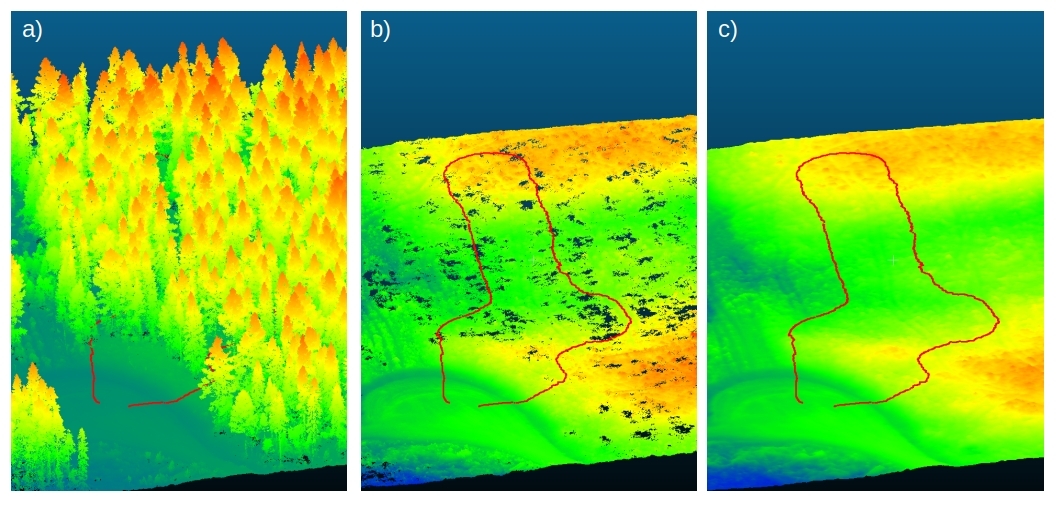}
    \caption{Visualization of the three steps from unclassified point cloud (a), via classified point cloud (b), to interpolated terrain model (c) for part of the Björsjö site, including our test circuit marked in red. Colors from blue to red reflect low to high elevation.}
    \label{fig:terrain_model}
\end{figure}

The laser scanning data is provided in .las format, which is a standard format for storing point cloud data.
GeoTIFF (.tif) files, based on ground classification of the laser scanning data, are provided for each site, with a resolution of between 0.04 and 0.1 m$^2$ per pixel.
Classified las-files containing only ground points are also provided in .las format.
All three steps from unclassified point cloud, via classified point cloud, to interpolated terrain model are visualized for part of the Björsjö site (Fig.~\ref{fig:terrain_model}).
Drone images with 60 m flight height are provided as orthomosaic in GeoTIFF format, with a resolution of 2.89 and 1.44 cm$^2$ per pixel for Märrviken and Björsjö respectively.
The photogrammetric point clouds based on drone footage are provided in .las format, split into one hectare tiles to reduce individual file size.
Each experimental site has its own set of photogrammetric point cloud files accompanied by a metadata .csv-file describing the extent of each tile, easily visualized in geographic information software such as QGIS.
Point density for the photogrammetric point clouds are about 5,000 and 3,000 points/m$^2$ for Märrviken and Björsjö respectively.
The small 3-D models created from mobile phone footage are provided in .obj and .laz format together with an .mp4 video, all standard formats for 3D models.

\subsection{Analysis tools}
Scripts for processing raw data and analysing curated synchronized data are provided in a separate repository available at the Swedish National Data Service (SND)~\cite{Lundback2025FORWARD}.
The scripts are written in Python and use mainly libraries such as NumPy and Matplotlib, along with some specialized libraries for reading and writing specific file formats.
Requirements and installation instructions together with usage examples are provided in the repository.
The scripts are structured in a modular way, allowing for easy extension and modification.
The main functionality is organized in a processing pipeline, where different modules can be combined to perform specific tasks on the data.
The pipeline can be constructed and run using a command line interface, optionally in combination with configuration files, with options to specify input and output directories, processing steps, and parameters.
The main steps in the processing pipeline are:
\begin{itemize}
    \item Loading raw data from different sources and formats, including appropriate synchronization of time series data.
    \item Conversion between different data types, for example turning text strings into integers to enable plotting.
    \item Extracting scenarios and episodes of interest from the synchronized data by selecting appropriate time intervals.
    \item Saving processed data in a separate npz file for easy loading in future analyses.
    \item Visualizing the data in different ways, including time series plots, histograms, and maps.
    \item Performing basic analyses, such as computing statistics and fitting models to the data.
\end{itemize}

Some of the most common data handling and analysis modules are listed in Table~\ref{tab:analysis_modules}, a complete list together with instructions on how to construct and run a number of useful commands is provided in the repository.

\begin{table}
    \small
    \centering
    \renewcommand{\arraystretch}{1.25}
    \caption{Some of the most common data handling and analysis modules provided in the analysis scripts repository.}
    \begin{tabular}{|p{3cm}|p{10cm}|}
    \hline
    Module & Description \\
    \hline
    \code{LoadIMU} & Load IMU data from a specified list of files and synchronize with machine time series data \\
    \code{LoadVibration} & Load vibration data from a specified list of files and synchronize with machine time series data \\
    \code{LoadGeoTif} & Load terrain data from a geotif file for subsequent plotting \\
    \code{TimelineGraphical} & Show the interactive plot of loaded time series data and video files \\
    \code{TimesToPick} & Specify start and end times for data extraction \\
    \code{Pick} & Perform the data extraction based on specified times \\
    \code{Save} & Save the extracted data in a npz file \\
    \code{PlotAll} & Plot all loaded variables in the time series data with indexed time on horizontal axis \\
    \code{Histogram} & Plot histograms of all loaded variables \\
    \code{ColorPath} & Plot all loaded variables as colored paths on the terrain map \\
    \code{SweRef} & Module to convert position data into the Swedish coordinate system Sweref99 TM \\
    \code{RemoveAcc} & Remove all but certain acceleration data variables from the dataset to avoid excessive number of plots \\
    \code{RemoveImu} & Remove all but certain IMU data variables from the dataset to avoid excessive number of plots \\
    \code{LinearRegression} & Fit linear regression models to a subset of the data, e.g., fuel consumption and driving speed \\
    \hline
    \end{tabular}
    \label{tab:analysis_modules}
\end{table}

Some of the preprocessing of terrain data, such as classification of point clouds and interpolation of terrain models, is performed with a more manual approach using R-scripts and existing tools like PDAL and CloudCompare.
Even though these steps are one-time processes, they are crucial for ensuring the quality and usability of the data in subsequent analyses, and for the sake of repeatability included and described in the repository.
In the following section, we present some example analyses that can be performed using the provided scripts.

\FloatBarrier

\section{Experimental Design, Materials and Methods}
\begin{figure}
    \centering
    \includegraphics[width=0.9\linewidth]{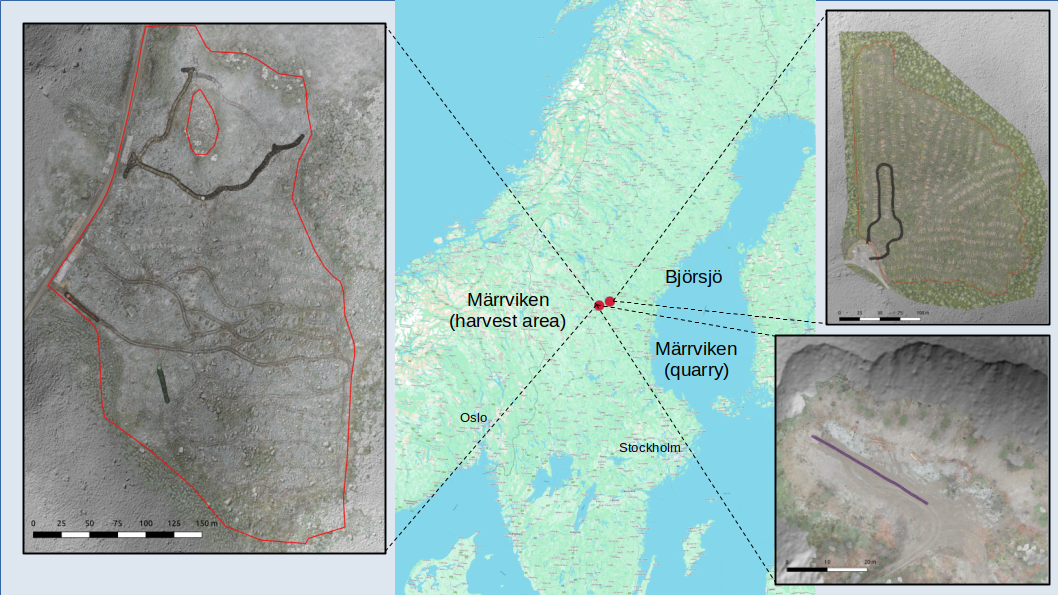}
    \caption{Overview of the testsites with terrain elevation maps obtained from the airborne laser scanning overlaid by orthomosaic photos from drone images and machine paths from the GNSS positioning data.}
    \label{fig:overview}
\end{figure}

Data origins from two field campaigns at two separate sites (Fig.~\ref{fig:overview}).
Märrviken is the first site with data gathered in October 2023 and very rough terrain with large boulders.
Björsjö is the second site with data gathered in September 2024 and terrain that is less rough but still includes instances of challenging boulders.
The temperatures were between -10 and 0\degree C with some overnight snowfall in Märrviken and around 15\degree C and sunny in Björsjö.

\subsection{Key machine dimensions and mass properties}
\begin{figure}[h]
    \centering
    \includegraphics[width=0.7\textwidth]{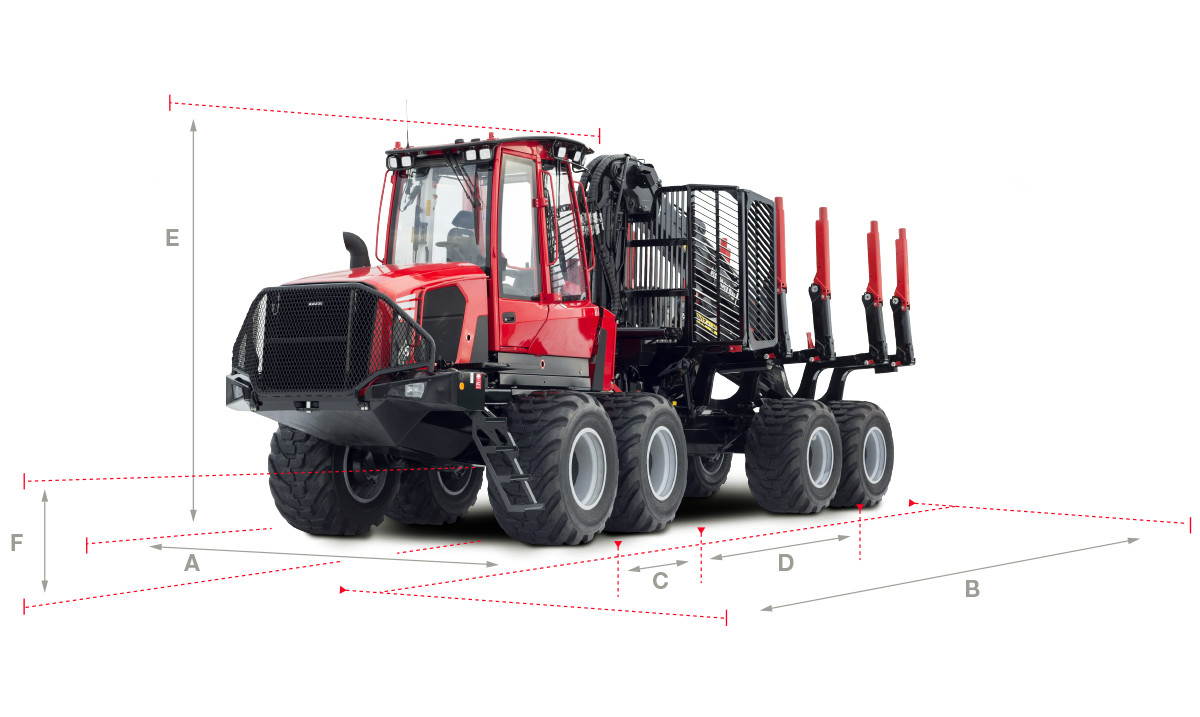}
    \caption{Key dimensions for the Komatsu forwarder.}
    \label{fig:895_dimensions}
\end{figure}

\begin{figure}[h]
    \centering
    \includegraphics[height=0.35\textwidth]{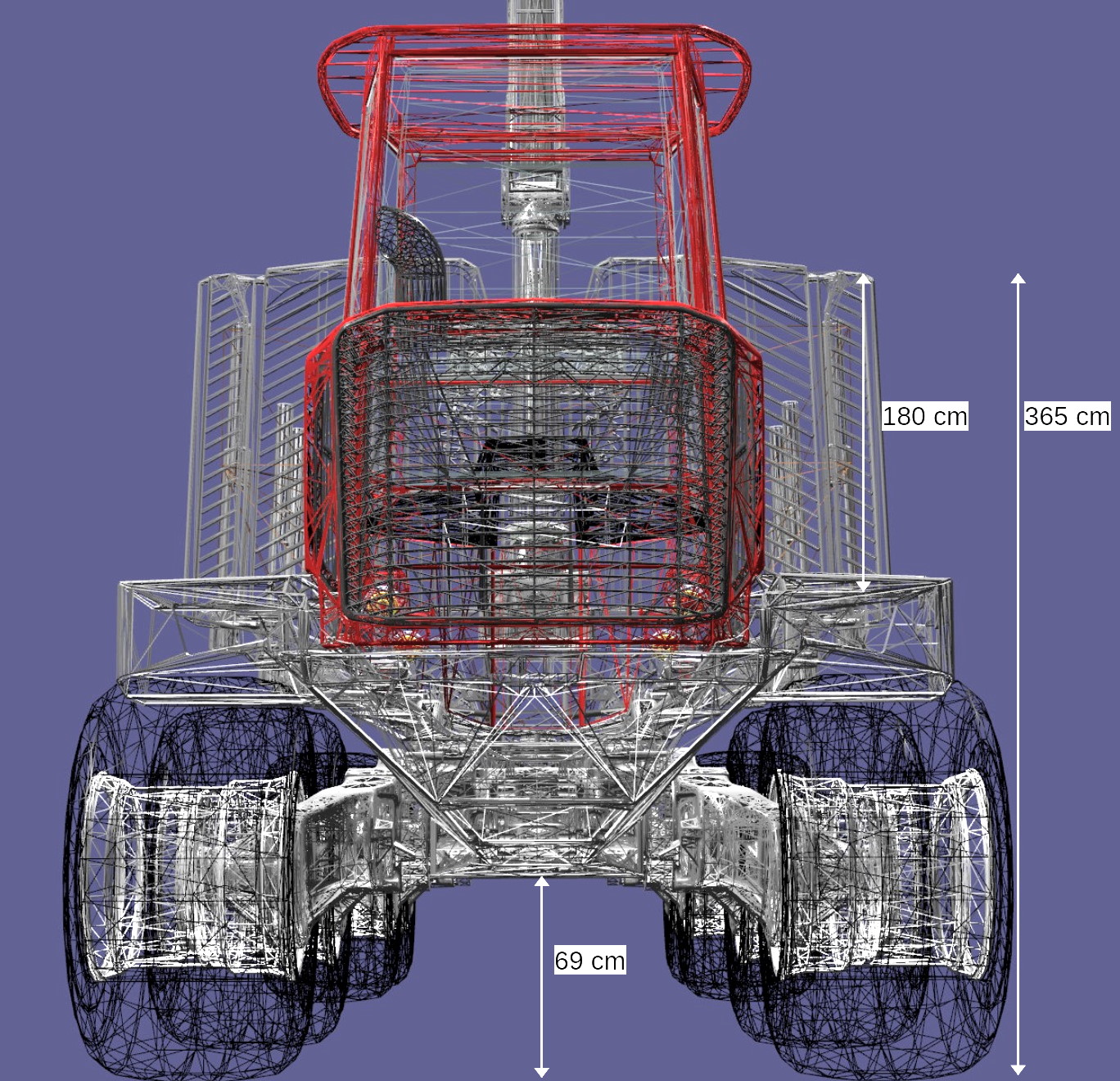}
    \includegraphics[height=0.35\textwidth]{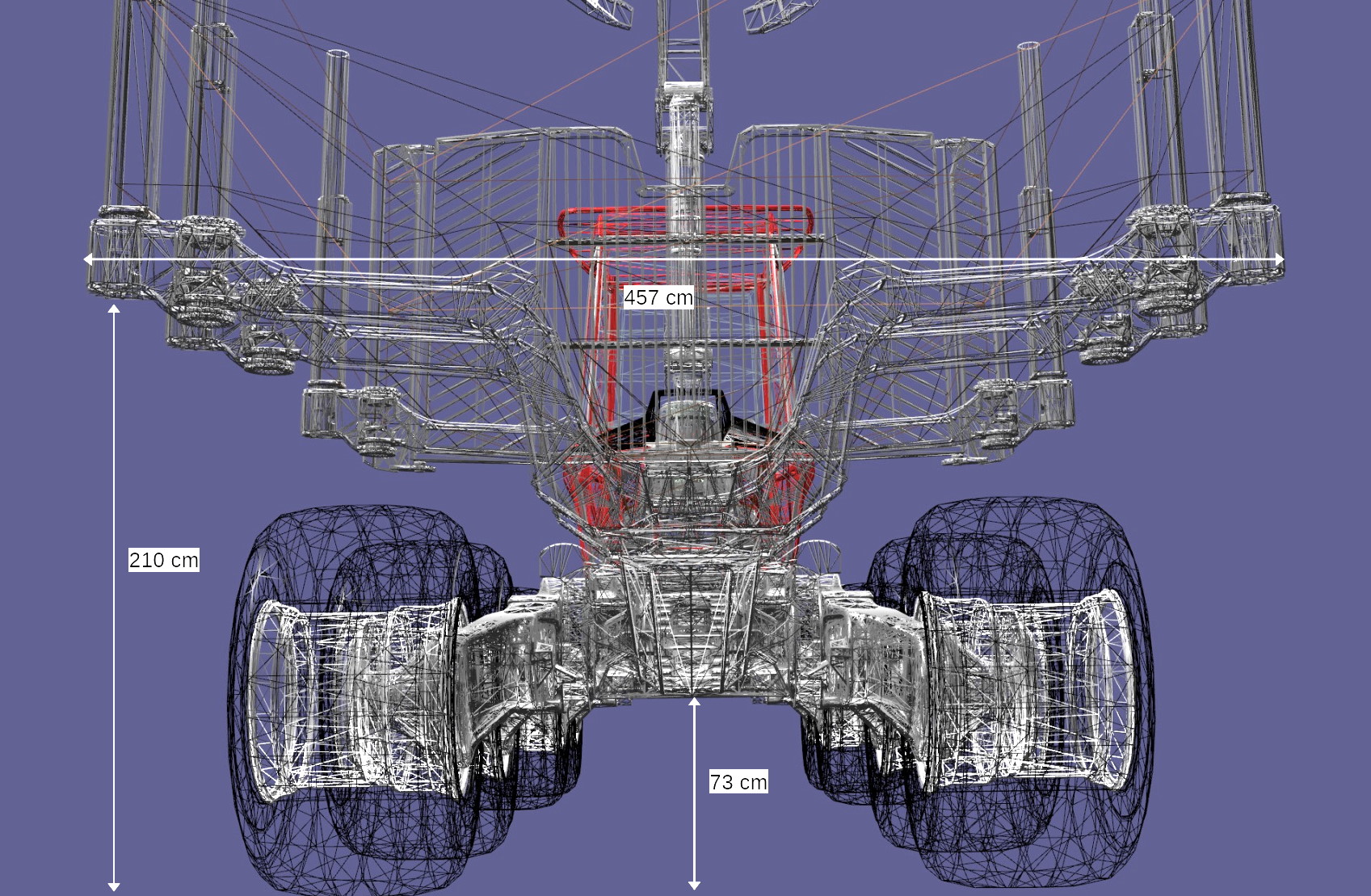}
    \includegraphics[height=0.352\textwidth]{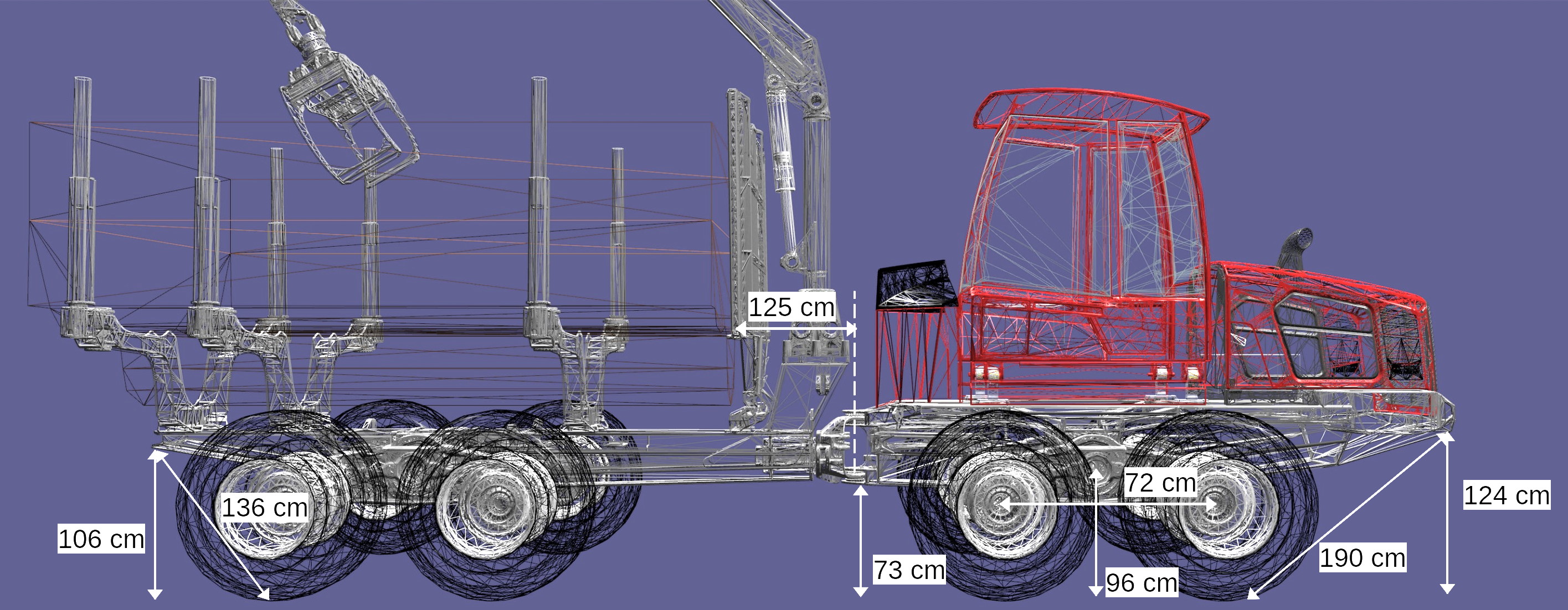}
    \caption{Measurements made during field test.}
    \label{fig:field_measurement}
\end{figure}

A Komatsu forwarder delivered new in 2023 was used in both campaigns/field tests.
The forwarder has the key properties listed in Table \ref{table:key} according to the manufacturer specification \cite{Komatsu895}, with reference to Fig.~\ref{fig:895_dimensions} and additional information.
In contrast to the specifications in Table \ref{table:key}, the total mass of the forwarder used in the field tests was 35,000 kg.
This includes chassis reinforcement, front blade, cabin with active damping, longer crane,
and tracks.
In most field tests, the forwarder was equipped with steel tracks on each wheeled bogie.
The tracks model was \texttt{Olofsfors ECO MAX 395 67 3100 OF}.
Their weight is 2,200 kg per pair, i.e., 1,100 kg per piece.
The driveline consists of a diesel engine, hydrostatic drive including pump and motor, and a mechanical drivetrain with gear reductions and boogie balancing.
The differential between front and rear axles is permanently locked while front and rear differentails between left and right are open by default, with the possibility for the operator to lock.
During the first field test (Märrviken, 2023-10-24) a number of key measurements were made.
These are shown in Fig.~\ref{fig:field_measurement}.

\begin{table}[h]
    \centering
    \caption{Key properties of a Komatsu forwarder from manufacturer specifications \cite{Komatsu895}.}
    \label{table:key}
    \begin{tabular}{ |l|c|r|}
     \hline
     Property & Unit & Value \\
     \hline
     Width (A) & mm & 3,160\\
     Length (B) & mm & 10,790\\
     Front axle to articulation joint (C) & mm & 2,000\\
     Rear axle to articulation joint (D) & mm & 3,900\\
     Articulation joint maximum slew angle & degrees & 42\\
     Articulation joint maximum roll angle & degrees & 18\\
     Ground clearance (F) & mm & 790\\
     Wheel radius & mm & 742\\
     Weight & kg & 23,600\\
     Max load & kg & 20,000\\
     \hline
    \end{tabular}
\end{table}

\FloatBarrier

\subsection{Terrain scans and data gathering}
Terrain data were collected before and during the experiments, using helicopter-borne laser scanning and drone footage-based photogrammetry respectively.
The target height of the laser scan with helicopter was 150 meters, resulting in point densities of approximately 1,500 points per square meter.
A Riegl Vux 120 sensor was used for the laser scanning and the time of scan is July 25-29 2022 and August 19 2023, for Märrviken and Björsjö sites respectively.

\begin{figure}
    \centering
    \includegraphics[height=0.35\linewidth]{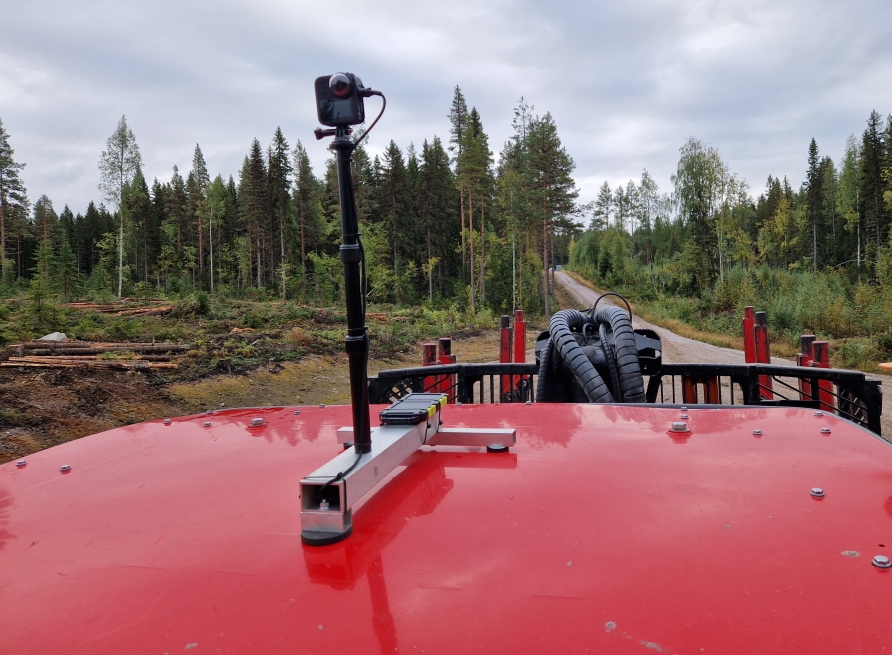}
    \includegraphics[height=0.35\linewidth]{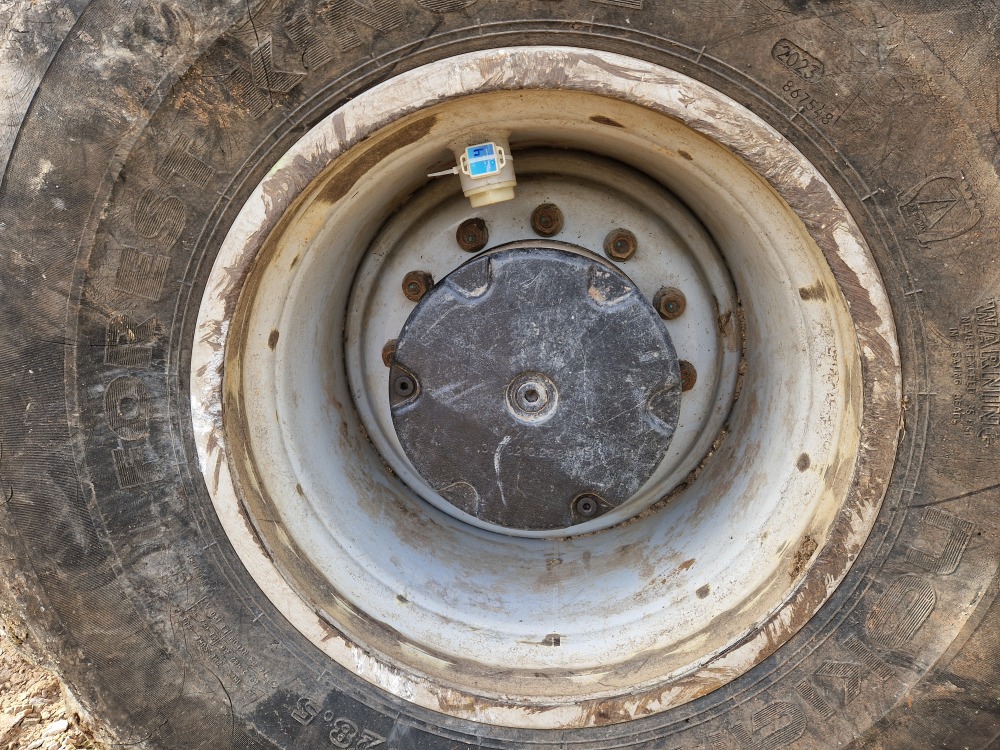}
    \caption{360\degree camera mounted on the forwarder cabin (left) and IMU mounted on the wheels (right).}
    \label{fig:360_cam_imu}
\end{figure}

During experiments and regular forwarding work at the Björsjö site, we mounted a 360\textdegree-camera on top of the forwarder cabin.
The 360\textdegree-camera records video and audio of the forwarder's surroundings (Fig.~\ref{fig:360_cam_imu}, left).
At the Märrviken site, a few key features were also recorded with a standard mobile phone camera, and converted into 3D models using the online service from Polycam (https://poly.cam).

The forwarder was operated by two professional forwarder operators, whose seat was equipped with a vibration sensor for parts of the experiments at the Björsjö site.
The vibration sensor used is a 3D accelerometer with a sampling frequency of 6,000 Hz.
For parts of the experiments at Björsjö, we also mounted a number of IMUs on the forwarder wheels, which record angular velocity at 5 Hz (Fig.~\ref{fig:360_cam_imu}, right).
The IMUs are synchronized with the machine data and the 360\textdegree-video, allowing for detailed analysis of the forwarder's movement and behavior during the experiments, with specific regard to wheel slip.

The camera data is time-synced using the timestamps provided in the metadata, with accuracy of 1 second.
The IMUs are synchronized by inducing and subsequently identifying characteristic events (rotation about a single axis, brief free-fall) at known time instants at the start and end of measurements.
Both synchronization methods achieve a temporal accuracy on the order of one second.
To enable post-processing and analysis such as synchronization of data streams and isolation of specific scenarios, start- and end-times of all experiments and other relevant episodes are manually logged in a field notebook (\texttt{MRV/BRJ\_log\_in\_Swedish.txt} in the dataset).

The machine data (including GNSS and CAN signals sampled at 5 Hz) was used as the primary temporal reference, and all other sensors were aligned and resampled to this.
The IMUs also sampled at 5 Hz.
However, as the vibration data were recorded at a substantially higher sampling frequency, this was aggregated to the machine-data timestamps using statistical summaries (mean, minimum, maximum, and standard deviation) computed over each time interval.

Because synchronization across modalities (video, vibration, and IMUs) relied on timestamps with 1 second resolution, and in some cases manual identification of synchronization events, the uncertainty is estimated to be on the order of 1 second.
This is sufficient for analyses at the scale of machine operations.
However, sub-second relationships between high-frequency signals (e.g., IMU and vibration measurements) should be interpreted with caution, unless further refinement of the synchronization is performed.

\subsection{Driving scenario naming convention}
Parts of the dataset is based on episodes of recorded machine data during controlled experiments at both the Märrviken and Björsjö sites. The following naming convention is used to organize that data:
\newline

\code{LOCATION\_SURFACE\_CONFIGURATION-DESCRIPTION\_1-DESCRIPTION\_2}

\vspace{3mm}\noindent
with: positional arguments for location and surface material using three-letter location code per site and one-word surface-material description;keyword arguments for systematically varied parameters; pairs of one-words, separated by underscore (\code{parameter\_value}); as principle for ordering, parameters which are more seldom varied are placed first; free text descriptions for characterising the test are separated by dash and multiple words combined with underscore. An example is:
\newline

\code{BRJ\_terrain\_tracks\_on\_load\_full\_inch\_020-test\_circuit-part\_1}
\newline

\vspace{3mm}\noindent
where \code{BRJ} stands for the Björsjö location, \code{terrain} indicates the surface material is  heterogeneous forest terrain in contrast to sand or gravel materials, \code{tracks\_on} means that steel tracks were mounted to the boogies, \code{load\_full} means full load of logs on the bunk of the machine, \code{inch\_020} means that the target speed during the experiment was 20\% of max speed on low gear (about 2.1 m/s), \code{test\_circuit} refers to the specific circuit that were used repeatedly at this specific location, and \code{part\_1} refers to a subset of the complete test circuit.

The following scenarios are extracted from the dataset and combined end-to-end, enabling e.g., iterative simulations on the whole set of scenarios.

\subsubsection{Märrviken - Driving on a flat smooth surface}
\begin{figure}
    \centering
    \includegraphics[width=0.9\linewidth]{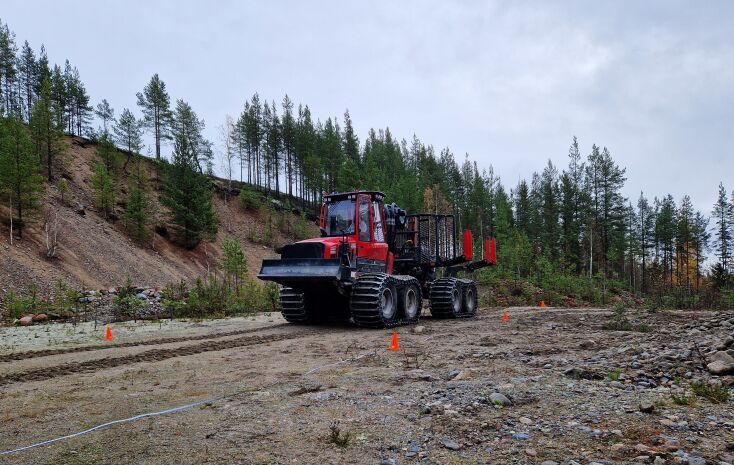}
    \includegraphics[width=0.45\linewidth]{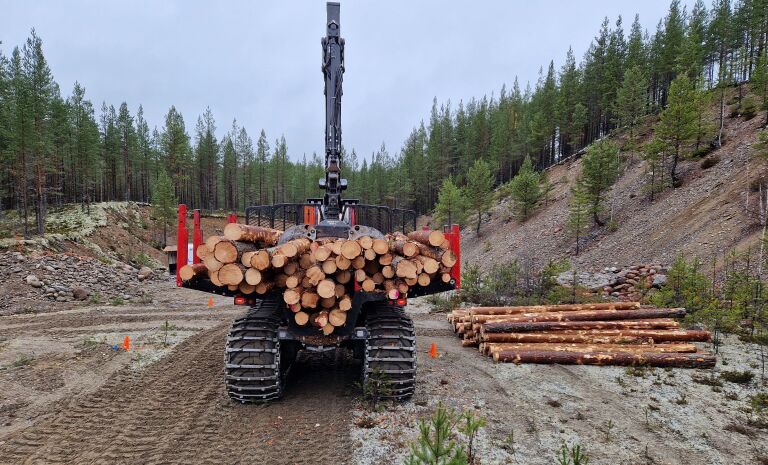}
    \includegraphics[width=0.45\linewidth]{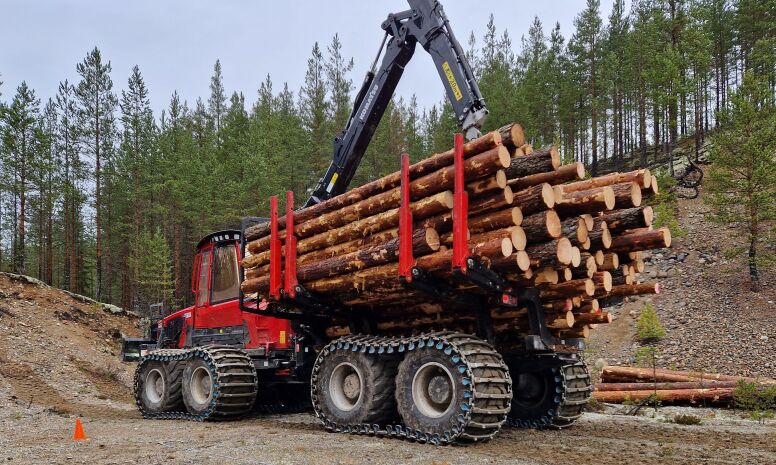}
    \caption{Forwarder during experiments on flat, sandy surface in old quarry. Empty, 10 tonnes, and 20 tonnes load weight.}
    \label{fig:quarry}
\end{figure}

\begin{figure}
    \centering
    \includegraphics[width=0.45\linewidth]{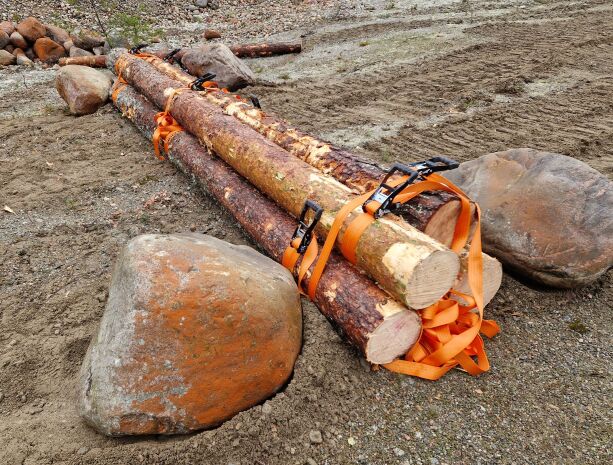}
    \includegraphics[width=0.45\linewidth]{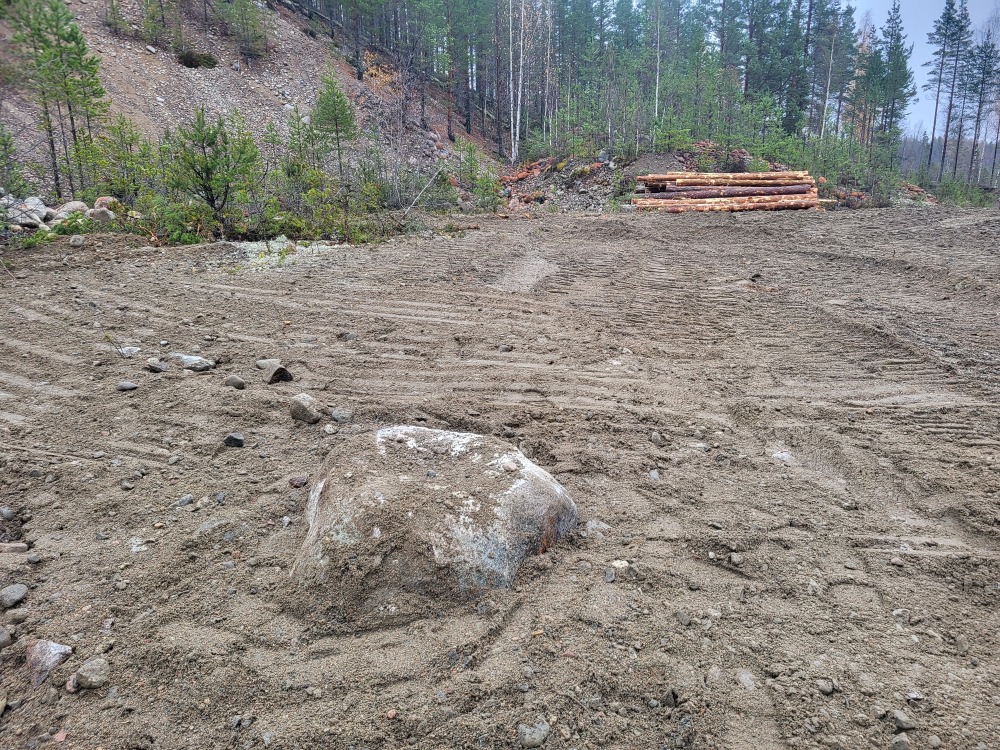}
    \caption{Log bundle and stone traversed by the forwarder in the quarry.}
    \label{fig:obstacles}
\end{figure}

We use an old quarry to facilitate a smooth and flat surface for repeated experiments, forming a baseline in the machine-dataset. Parameters adjusted in the experiments are target driving speed (10\%, 30\%, 50\%, 70\%, 90\%, and 100\% of full speed in terrain gear), load weight (0, 10, 20 tonnes) (Fig.~\ref{fig:quarry}). A bundle of logs and a dug down stone represent obstacles that are relatively easy to recreate in simulation (Fig.~\ref{fig:obstacles}). The obstacles are traversed at 30\% target speed with all three load weights.

\subsubsection{Märrviken - Driving uphill on gravel road}
\begin{figure}
    \centering
    \includegraphics[height=0.45\linewidth]{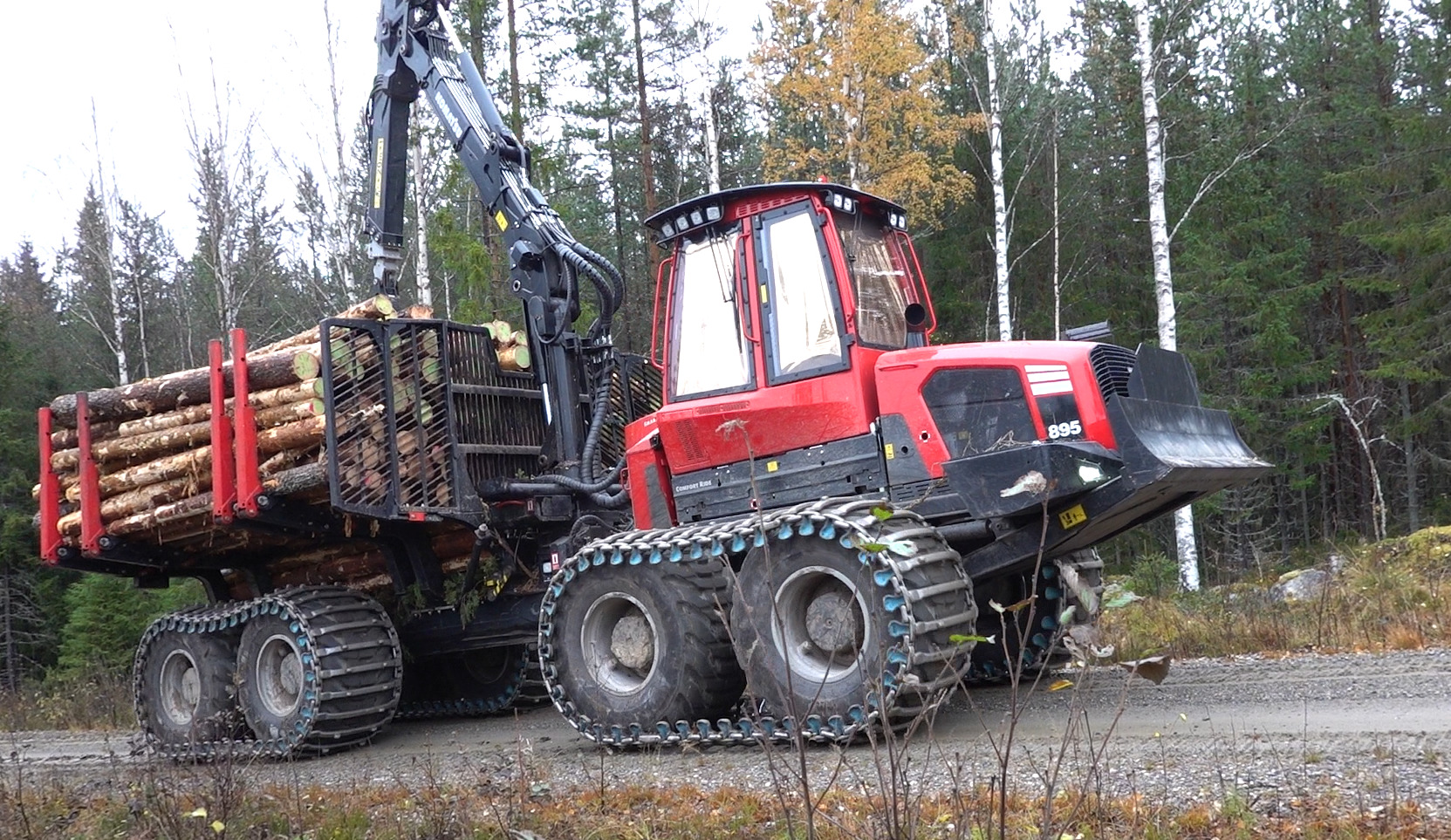}
    \includegraphics[height=0.45\linewidth]{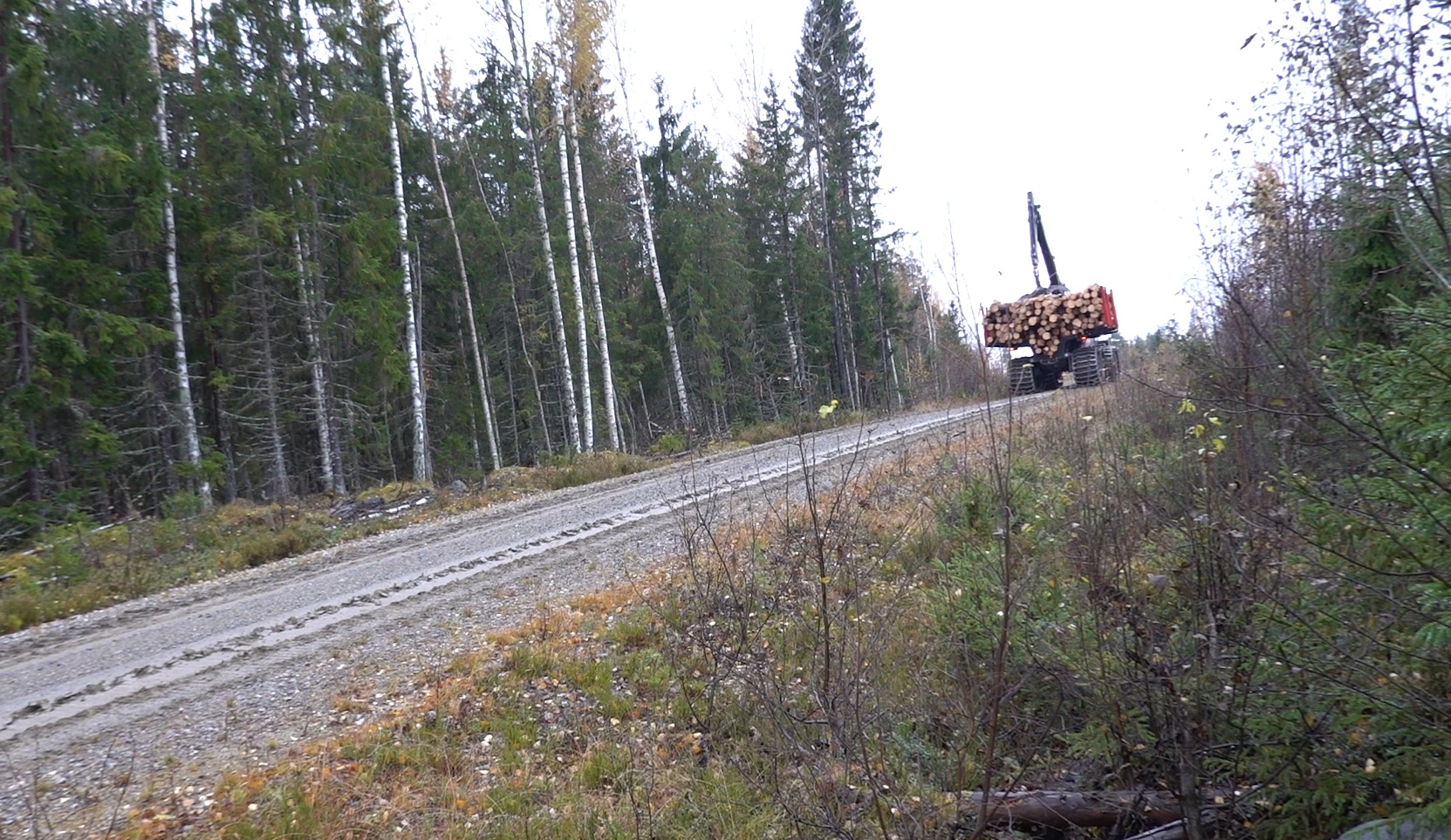}
    \caption{Forwarder driving uphill forestry road at the Märrviken site.}
    \label{fig:road_uphill}
\end{figure}

Target speed levels of 30\%, 60\%, and 100\% as well as all three load weights are also tested on an inclined forestry road, consisting of hard-packed gravel. The uphill slope is about 5\degree, see Fig.~\ref{fig:road_uphill}.

\subsubsection{Märrviken - Driving on flat terrain}
\begin{figure}
    \centering
    \includegraphics[width=0.9\linewidth]{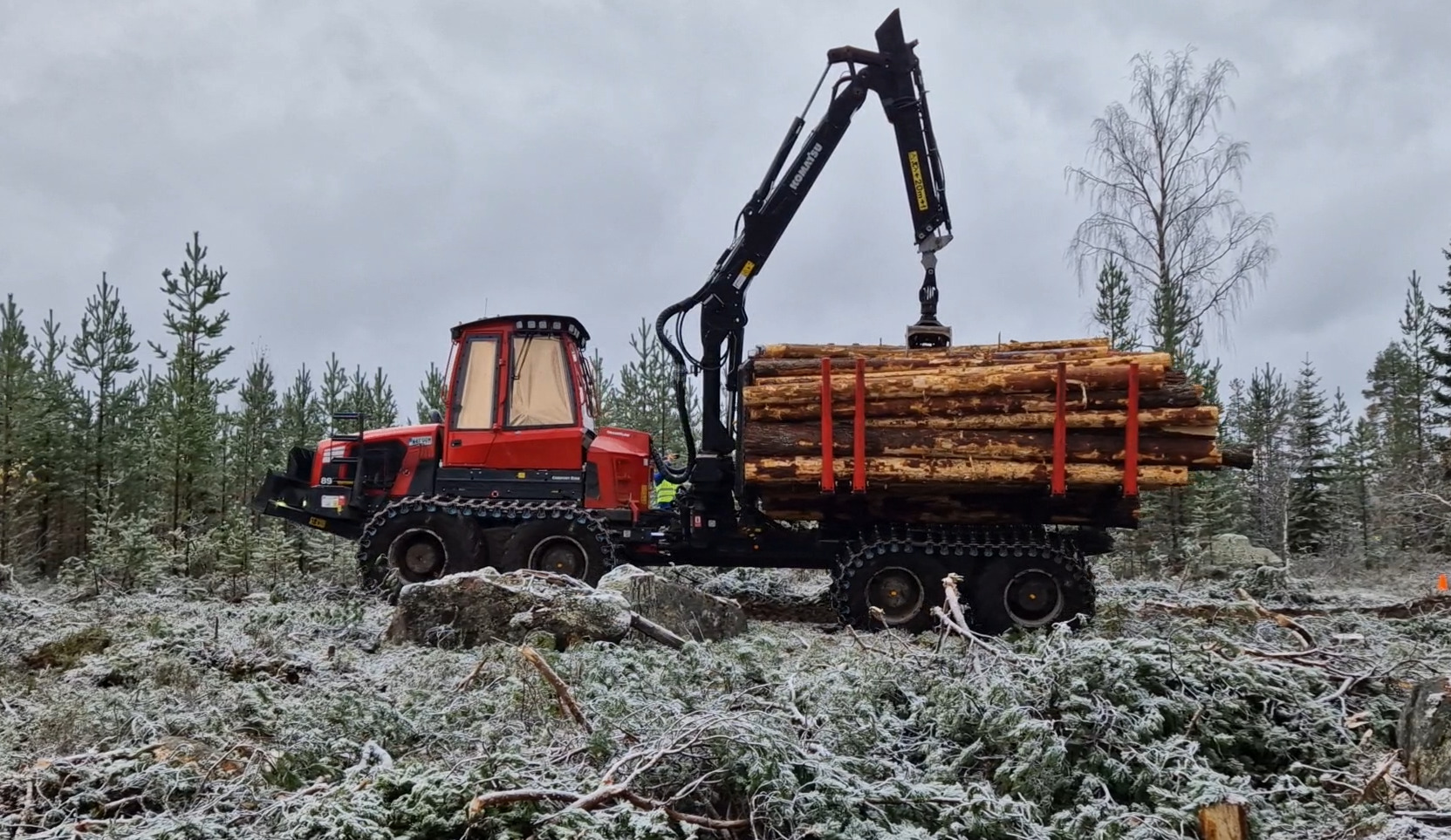}
    \caption{Forwarder on flat strip road at Märrviken with modest prevalence and size of obstacles.}
    \label{fig:striproad}
\end{figure}

On a flat part of a strip road we do similar tests. Target speeds of 10\%, 30\%, and 50\%, load weights of 0, 10, and 20 tonnes. Obstacles (stones) are present but not severe (Fig.~\ref{fig:striproad}).

\subsubsection{Märrviken - Driving in uphill terrain}
\begin{figure}
    \centering
    \includegraphics[width=0.9\linewidth]{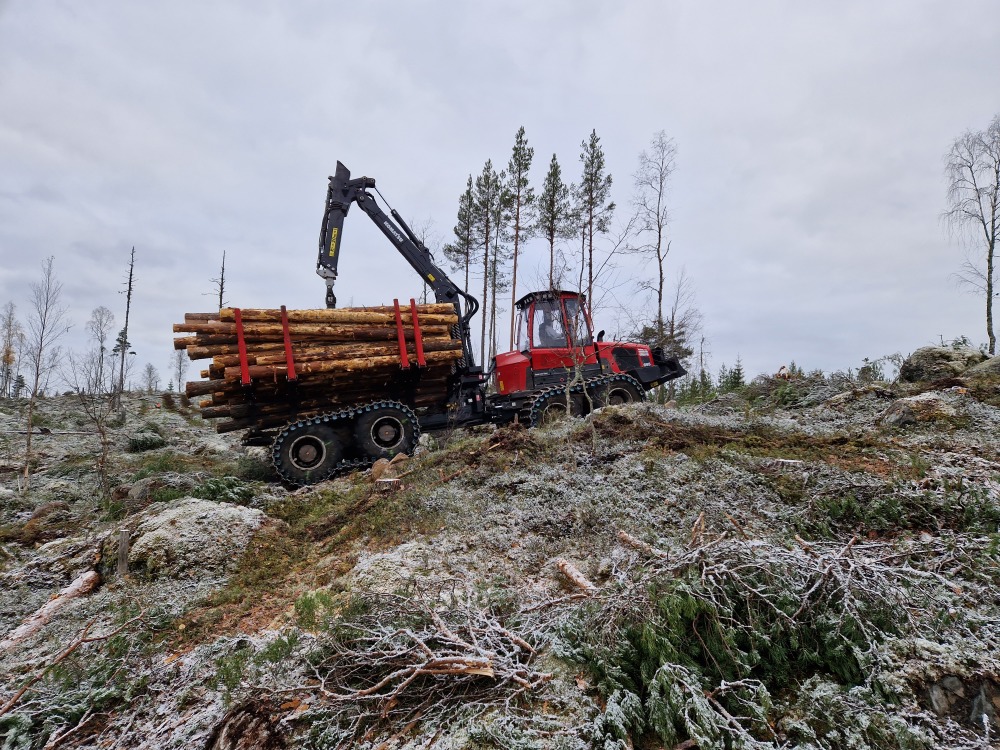}
    \caption{Forwarder on uphill strip road at Märrviken with modest prevalence and size of obstacles.}
    \label{fig:terrain_uphill}
\end{figure}

On a moderately inclined (about 5\degree) part of a strip road in the terrain we let the forwarder drive uphill with empty load at target speeds 10\%, 30\%, and 50\%, as well as with 20 tonnes load weight at target speeds 20\%, 30\%, 40\%, 60\%, and 80\%. Obstacles are present but not affecting possible driving speed (Fig.~\ref{fig:terrain_uphill}).

\subsubsection{Märrviken - One minute cycles in terrain}
\begin{figure}
    \centering
    \includegraphics[width=0.9\linewidth]{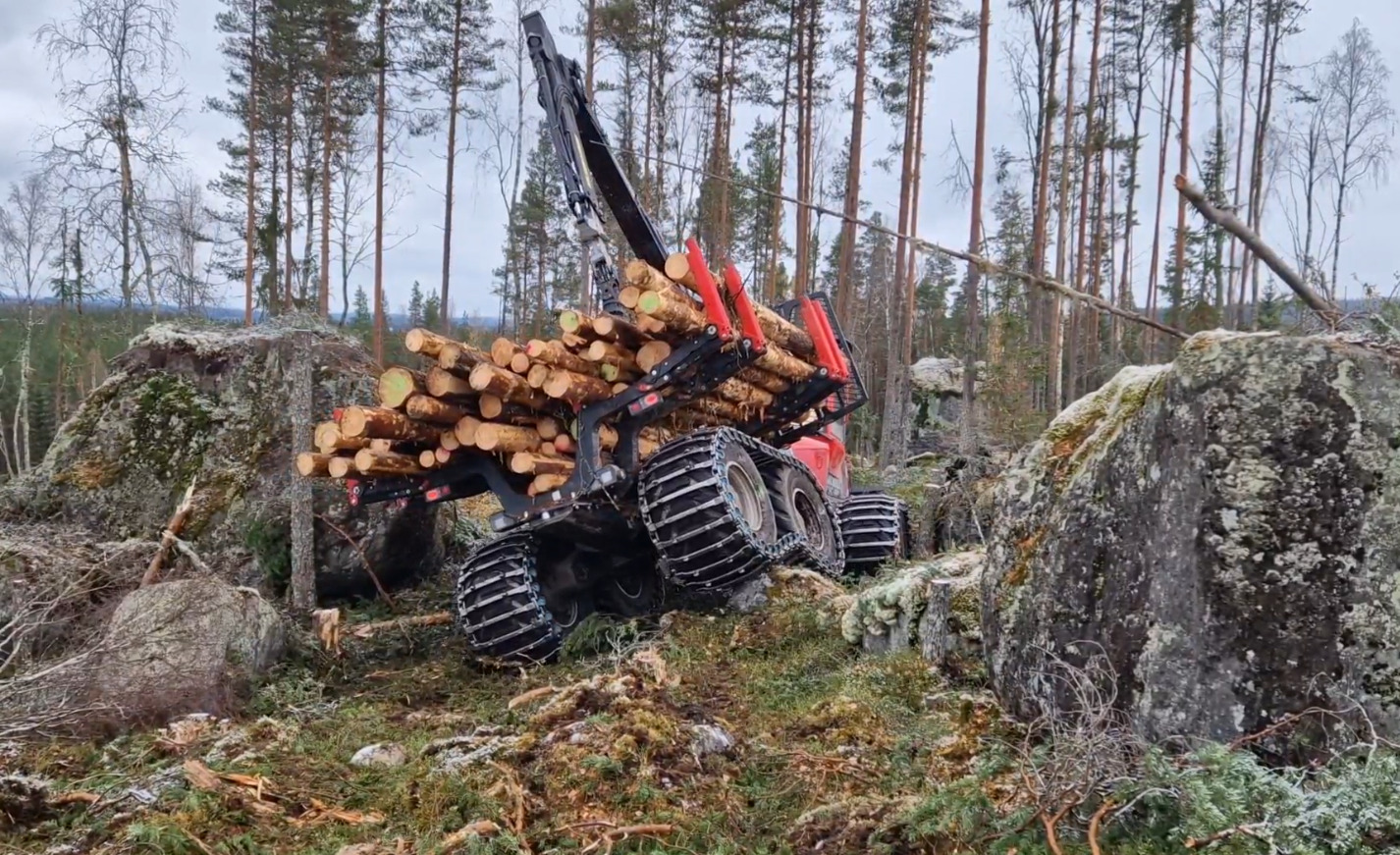}
    \caption{Forwarder on difficult strip road at Märrviken with rich prevalence and size of obstacles.}
    \label{fig:cycle}
\end{figure}

On a partly very rough path with big boulders, the forwarder travels at target speed 30\% and empty as well as 10 tons load weight. The sequences consists of downhill and uphill driving in very rocky and difficult terrain, example in Fig.~\ref{fig:cycle}.

\subsubsection{Märrviken - Driving steep uphill}
\begin{figure}
    \centering
    \includegraphics[width=0.45\linewidth]{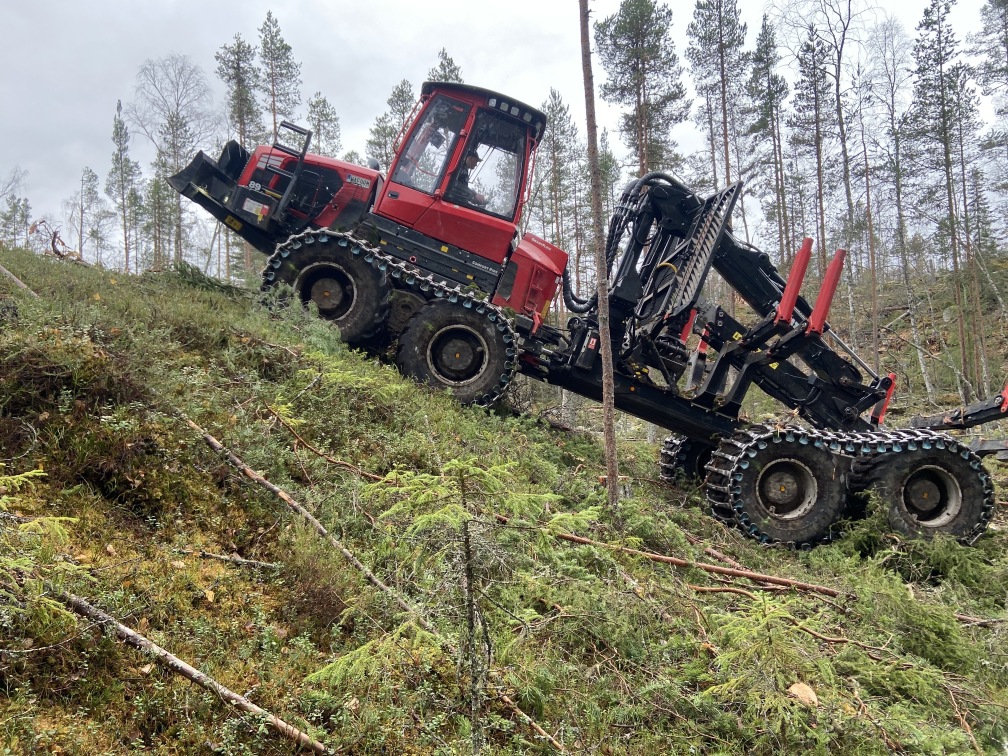}
    \includegraphics[width=0.45\linewidth]{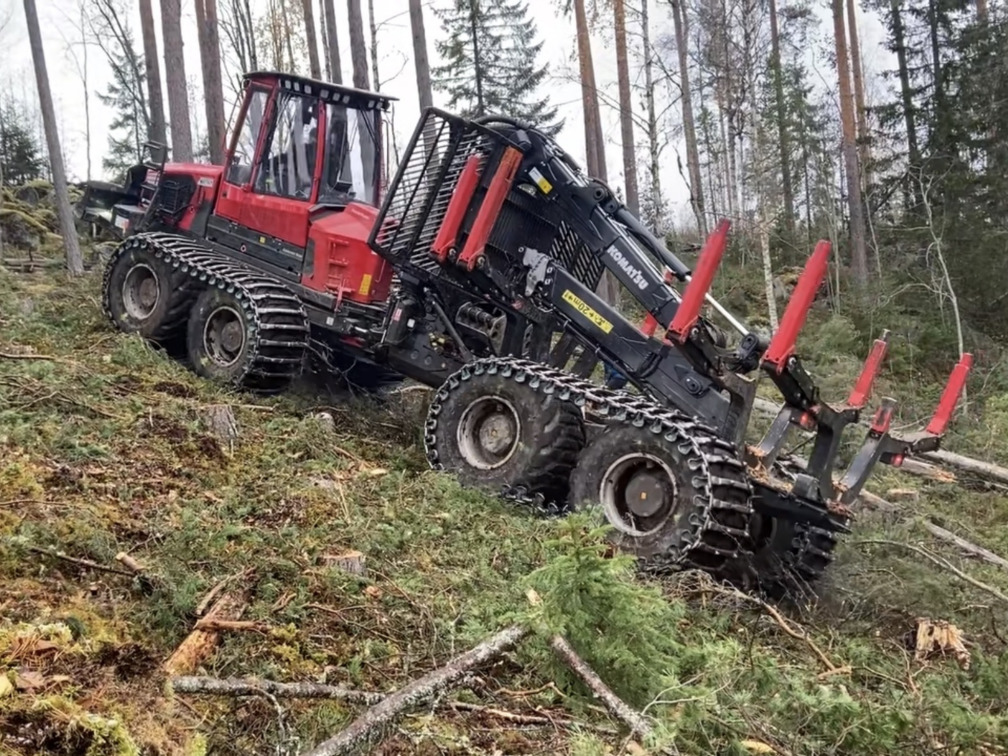}
    \caption{Forwarder driving steep uphill at Märrviken, smooth ground (left) and rough (right).}
    \label{fig:steep}
\end{figure}

In search of the limitations of the forwarder, we try two very steep uphill parts of the harvested area. The first quite smooth and the second rough, with stones and stumps (Fig.~\ref{fig:steep}). The slope was about 13\degree.

\subsubsection{Björsjö - Driving on gravel road}
\begin{figure}
    \centering
    \includegraphics[angle=270,width=0.45\linewidth]{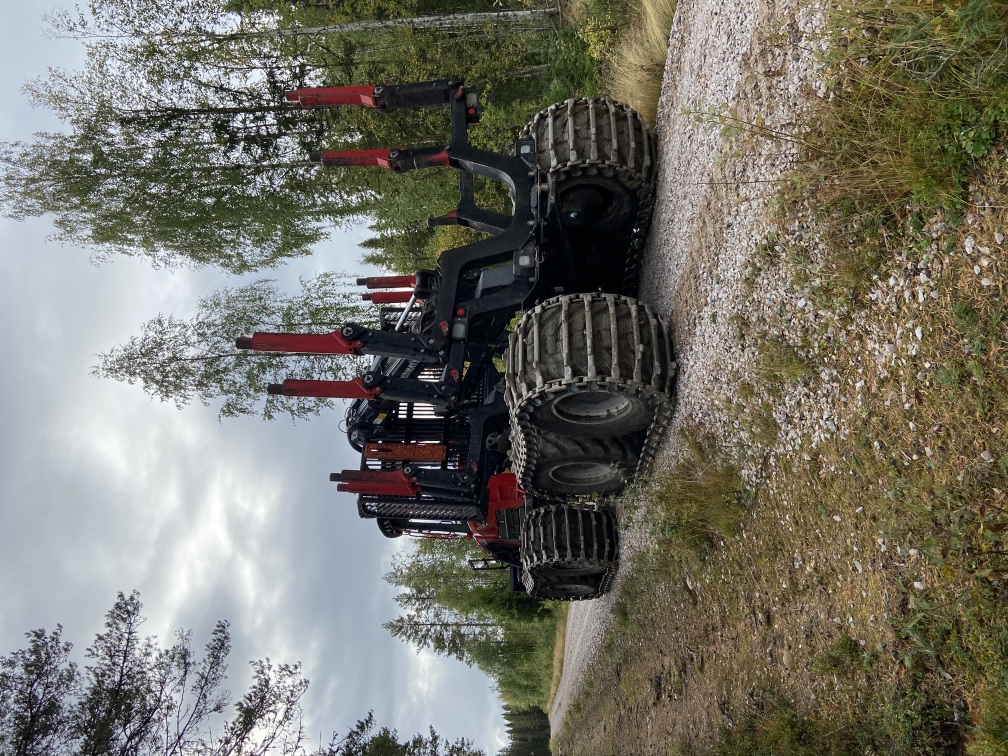}
    \includegraphics[angle=270,width=0.45\linewidth]{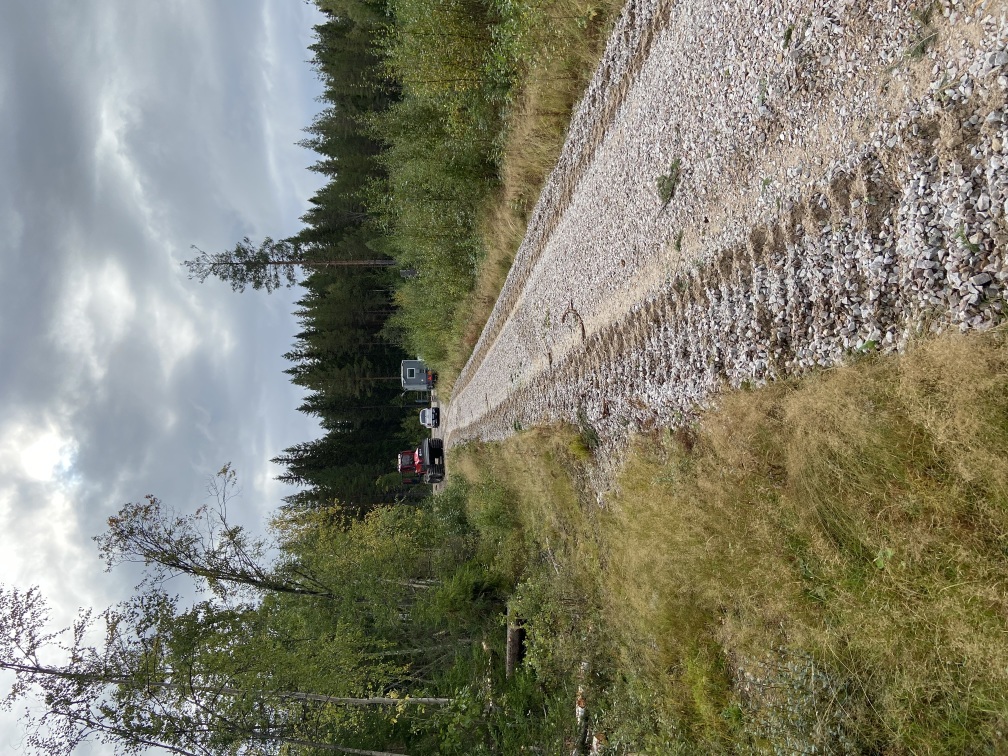}
    \caption{Forwarder on forestry road in Björsjö.}
    \label{fig:bj_road}
\end{figure}

On a nearly flat gravel road with fresh, quite soft gravel we drive with 30\%, 60\%, and 100\% of max speed and with empty as well as full load (Fig.~\ref{fig:bj_road}). The road section is 300 meters long with a very slight hill in the middle. The turnaround at the far end is deleted from the scenarios, leading to each combination of speed and load consisting of two time intervals. In the free text description part of our scenario-naming nomenclature we mark the two time intervals as \code{away} and \code{back}.

\subsubsection{Björsjö - Terrain test circuit}
\begin{figure}
    \centering
    \includegraphics[height=0.60\linewidth]{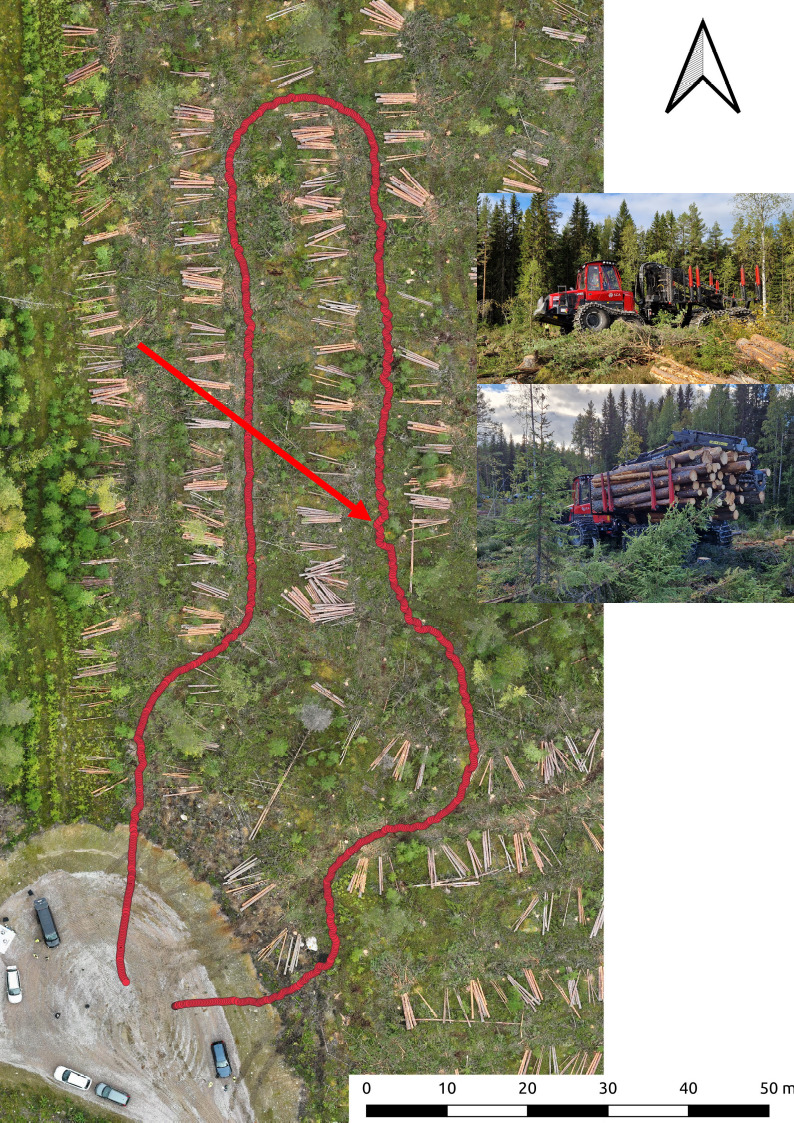}
    \includegraphics[height=0.61\linewidth]{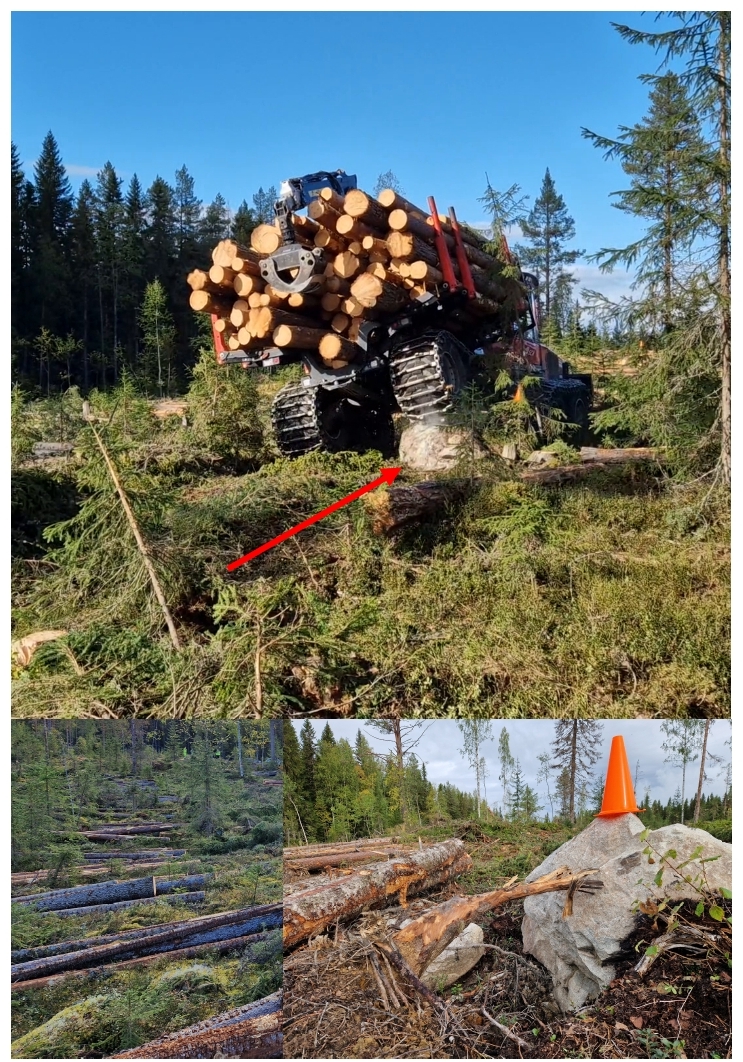}
    \caption{The terrain test circuit in red (left), 90 cm stone by the red arrow and 50 cm stone as inset in lower right corner (right).}
    \label{fig:bj_circuit}
\end{figure}

Along a test circuit following strip roads and some areas between strip roads, the forwarder traveled at target speeds 20\%, 30\%, and 40\% with empty and full load weight (Fig.~\ref{fig:bj_circuit}).
The test circuit is smooth in general with light incline on the way out and decline on the way back on a parallel strip road. At a few places along the circuit the forwarder traverses large stones with the wheels, the largest 90 cm high (Fig.~\ref{fig:bj_circuit}).

\FloatBarrier

\subsection{Annotation of work elements from video}
During the three days of field trails at the Björsjö-site, the forwarder is also operated for 18 hours in regular wood extraction work with the 360\textdegree-camera mounted on the cabin (Fig.~\ref{fig:360_cam_imu}).
This work is annotated into 8 different work elements following established ontology~\cite{bjorheden_international_2000}, see Table~\ref{tab:work_elements}.
All annotation were made by one single PhD-student during a few weeks and according to instructions from the research group.
No formal validation of the annotation were made, although the resulting plots and analyses seems sound.
Validation can be done by any user based on the annotation file and the lightweight 360\textdegree-videos included in the dataset.
The main objective of the annotation is not to investigate productivity on this specific harvesting site, but to complement the other data with information of work element, e.g., differing between driving with empty or fully loaded bunk (35 vs 55 metric tonnes total machine weight).
By analysing annotated work elements together with machine-, position-, and terrain data we also see opportunities to advance automatic work element detection (AWED) in forwarding work.
The work elements are defined as follows:

\renewenvironment{description}[1][0pt]
  {\list{}{\labelwidth=0pt \leftmargin=#1
   \let\makelabel\descriptionlabel}}
  {\endlist}
\begin{description}[2mm]
    \item[Driving empty:] Driving without load from the landing to the place of the first log pick up.
    \item[Loading:] Picking up logs from the ground and placing them on the bunk. This work element is prioritized over \emph{Driving while loading} if both occur simultaneously.
    \item[Driving while loading:] Driving between log piles without simultaneously loading logs.
    \item[Driving loaded:] Driving with a full bunk of logs from the place of the last log pick up to the landing.
    \item[Unloading:] Picking up logs from the bunk and stacking them on the landing. This work element is prioritized over \emph{Driving while unloading} if both occur simultaneously.
    \item[Driving while unloading:] Driving between or along log stacks without simultaneously unloading logs.
    \item[Short delays:] Short stops for e.g., planning the work, adjusting the machine, or answering the phone. Typically a few seconds but up to five minutes.
    \item[Other time:] Time not spent on any of the above work elements.
\end{description}

\begin{figure}
    \centering
    \includegraphics[width=0.9\linewidth]{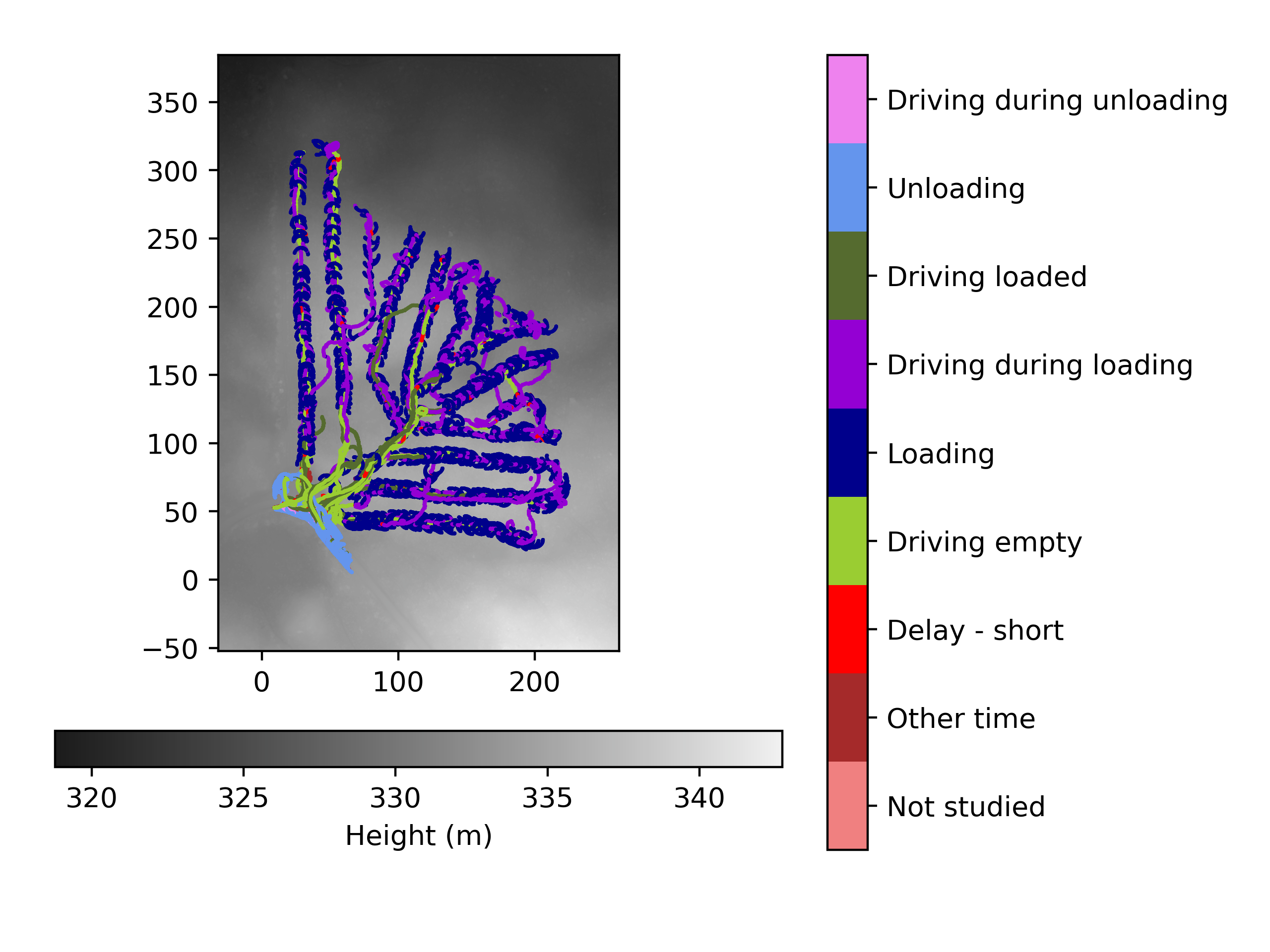}
    \caption{Map of the Björsjö-site with work elements during all regular work plotted on a height map based on the laser scanning. Crane movement from the crane signal data and GNSS-recordings are shown while loading and unloading, whereas forwarder movement by GNSS-recordings are shown in the rest of the work elements. Local northing and easting are shown in meters.}
    \label{fig:crane_all_work}
\end{figure}

\begin{figure}
    \centering
    \includegraphics[width=0.9\linewidth]{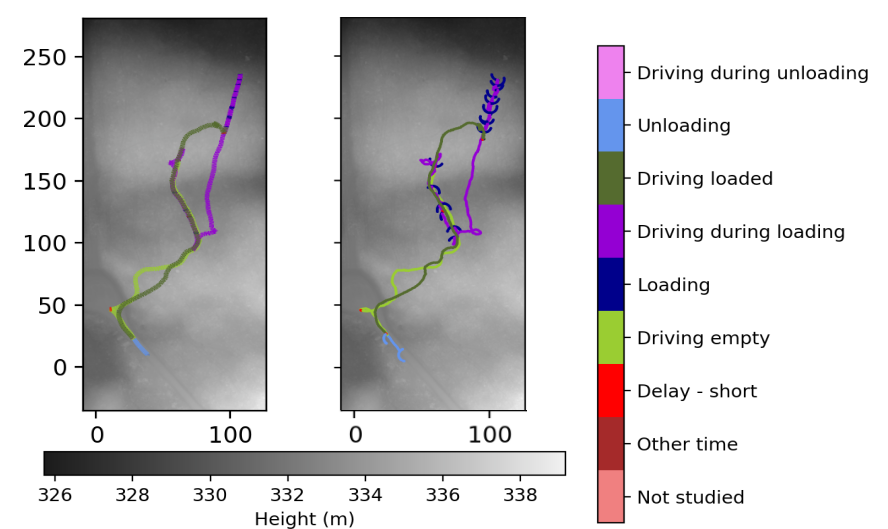}
    \caption{GNSS-recordings from one load cycle during regular work at the Björsjö-site, without (left) and with (right) crane movement data. Local northing and easting are shown in meters.}
    \label{fig:crane_cycle}
\end{figure}

The work element annotations are synchronized to the 360\textdegree-video material and incorporated in the 5 Hz time series machine data.
The active work element can be visualized on the map of the harvesting site using the annotation, exemplified in Fig.~\ref{fig:crane_all_work}.
By utilizing the crane signal data, we can also see when the crane is used, and for how long, during the different work elements.
Furthermore, individual load cycles are extracted from the data and used for further visualization and analysis (Fig.~\ref{fig:crane_cycle}).
This is done similarly to the extraction of experiment scenarios, by picking appropriate time intervals from the dataset with the tools provided in the analysis scripts.

\begin{table}
    \centering
    \caption{Time spent on different work elements during the video recorded part of the field trails at the Björsjö-site.}
    \begin{tabular}{|l|c|}
    \hline
    Work element & Time [h] \\
    \hline
    Driving empty & 1.8 \\
    Loading & 7.5 \\
    Driving while loading & 2.3 \\
    Driving loaded & 1.4 \\
    Unloading & 3.3 \\
    Driving while unloading & 0.05 \\
    Short delays & 1.1 \\
    Other time & 0.9 \\
    Total time annotated & 18.3 \\
    \hline
    \end{tabular}
    \label{tab:work_elements}
\end{table}

\FloatBarrier

\section{Applications}
To give more insight into the dataset and exemplify applications, we present two conceivable use-cases of the dataset and a visualization of selected StanForD production data, all from the Björsjö site.
The first example is a visualization of the fuel consumption during regular work, and the second example demonstrates the vibration data from the forwarder chassis and the operator seat. All examples are from the Björsjö site.

\subsection{Fuel consumption}
As with the other machine variables, the fuel consumption is recorded at 5Hz and is measured in two ways: instantaneous fuel consumption in liters per hour and accumulated fuel consumption in half liter steps.
The machine was not equipped with fuel flow sensors, so the manufacturer models fuel consumption using the opening time, temperature, and pressure of the fuel injector.
The details about this model are not disclosed by the manufacturer.
They state that the accumulated fuel consumption is typically more reliable; the reliability does, however, come at the cost of coarser resolution, as the accumulated fuel consumption is measured in steps of 0.5 liters.
The difference in accuracy between accumulated and instantaneous fuel consumption is not quantified by the manufacturer.

\begin{figure}
    \centering
    \includegraphics[width=0.9\linewidth]{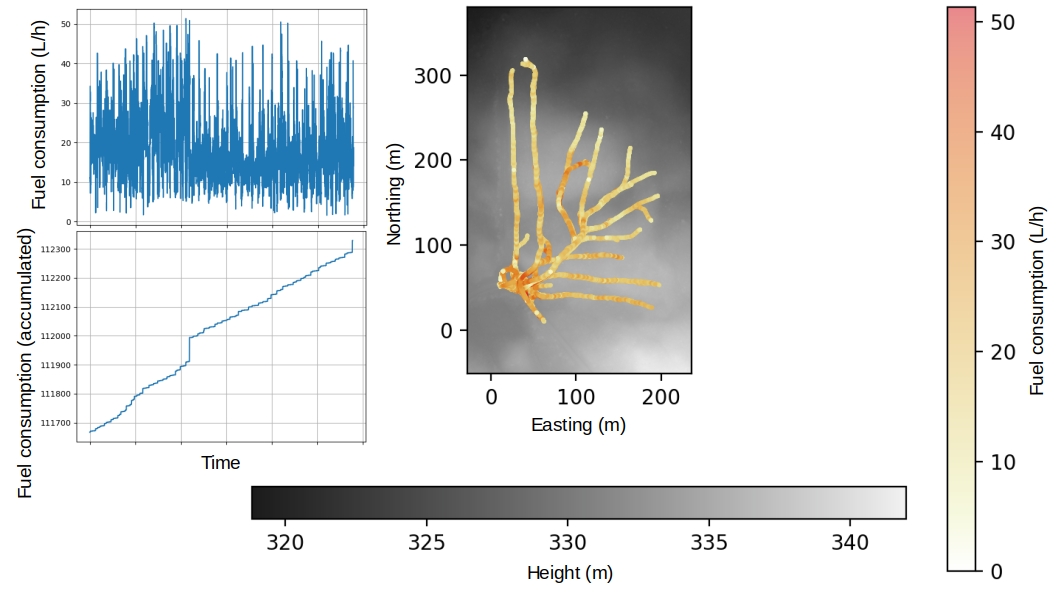}
    \caption{Fuel consumption during driving at the Björsjö-site. Instantaneous (a) and accumulated (b) fuel consumption as a time series, and along the machine paths with high values in red and low values in white (c). Local northing and easting are shown in meters.}
    \label{fig:fuel_consumption}
\end{figure}

\begin{figure}
    \centering
    \includegraphics[width=0.9\linewidth]{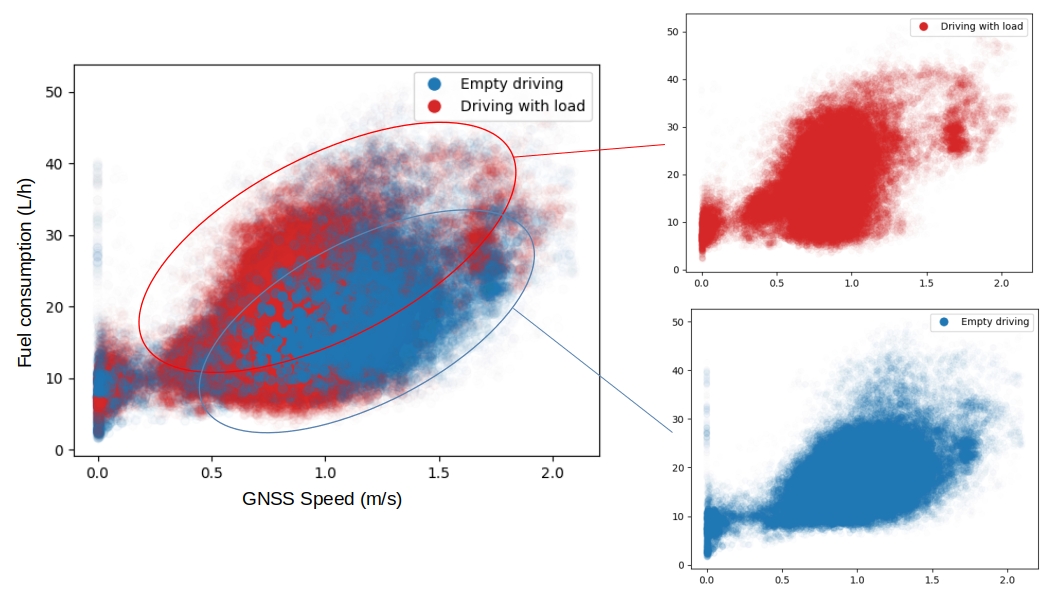}
    \caption{Instantaneous fuel consumption plotted against driving speed during regular work at the Björsjö-site. The data is filtered so that only work elements \emph{Empty driving} (blue) and \emph{Driving with (full) load} (red) are represented. The total number of observations is 57 802, corresponding to about 3 hours of driving.}
    \label{fuel_speed_driving}
\end{figure}

Focusing on instantaneous fuel consumption, we can plot fuel consumption against the driving speed recorded by the GNSS receiver, and confirm that higher speeds correlate with higher fuel consumption and that the speed-dependent increase is slightly more pronounced with a fully loaded bunk (Fig.~\ref{fuel_speed_driving}).
There is a big overlap in fuel consumption between the two work elements 'Empty driving' and 'Driving with (full) load', indicating several confounding factors.

\begin{figure}
    \centering
    \includegraphics[width=0.9\linewidth]{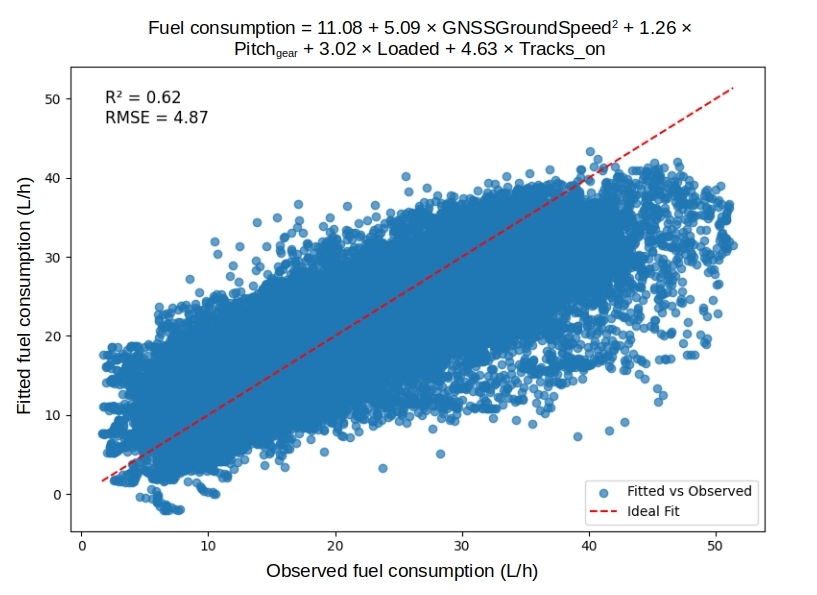}
    \caption{Linear regression model of instantaneous fuel consumption during \emph{'Empty driving'} and \emph{'Driving with load} at the Björsjö-site. The model is fitted to 57,802 observations, corresponding to about 3 hours of driving. The $R^2$ is 0.62 and the RMSE is 4.87 l/h and the plot show the model predictions against the actual fuel consumption. The red line is the ideal case where the model predicts the actual fuel consumption perfectly.}
    \label{fig:fuel_regression}
\end{figure}

We demonstrate modeling the fuel consumption using linear regression with the input variables driving speed, pitch angle, and booleans for full load and tracks on.
The resulting model has an $R^2$ of 0.62 and a root mean square error (RMSE) of 4.87 l/h.
This suggests that a fuel consumption model for driving in rough terrain needs to account for more factors than speed, slope, and load weight (Fig.~\ref{fig:fuel_regression}).
An interesting observation is that, in this dataset, the use of steel tracks seems to affect the fuel consumption more than the load weight.

\FloatBarrier

\subsection{Vibration data} \label{sec:vibration}
For a subset of the data from the Björsjö site, we have vibration data recorded from the operator seat.
The vibration data is recorded at 6,000 Hz and the vibration dose value (VDV), whole-body vibration exposure A(8), and daily equivalent static compression dose S$_{ed}$ is computed for a number of test scenarios and regular forwarding work (Table~\ref{tab:vibration_measures}).
Vibration dose value (VDV) and whole body vibration exposure A(8) is computed from the raw data using the ISO 2631-1 standard \cite{ISO_2631-1_1997}, while a daily equivalent static compression dose (S$_{ed}$) value is computed according to ISO 2631-5 \cite{ISO_2631-5_2018}.
For a stitched scenario containing 11 hours and 20 minutes of regular forwarding work, the VDV is 20.0 m/s$^{1.75}$ and A(8) is 0.66 m/s$^2$.
We also synchronize the data from the vibration sensor with the machine-data by downsampling it to 5 Hz, using metrics such as maximum, minimum, average, and standard deviation.
The downsampled vibration data lack sufficient resolution to be used for computation of human vibration exposure measures, but it is useful for exploring similarities between the vibration data and the machine data.

\begin{table}
    \centering
    \small
    \renewcommand{\arraystretch}{1.25}
    \caption{Vibration dose value (VDV), daily vibration exposure A(8), and daily equivalent static compression dose (S$_{ed}$) for different scenarios at the Björsjö-site.}
    \begin{tabular}{|l|c|c|c|c|}
    \hline
    Scenario & Time duration & S$_{ed}$ & VDV & A(8) \\
    \hline
    \code{BRJ\_terrain\_tracks\_on\_load\_empty\_inch\_030-test\_circuit} & 00:06:20 & 1.1445 & 34.44 & 0.85 \\
    \code{BRJ\_terrain\_tracks\_on\_load\_empty\_inch\_040-test\_circuit} & 00:04:30 & 1.2116 & 59.19 & 1.35 \\
    \code{BRJ\_terrain\_tracks\_on\_load\_empty\_inch\_020-test\_circuit} & 00:10:59 & 1.0442 & 17.46 & 0.32 \\
    \code{BRJ\_gravel\_tracks\_on\_load\_full\_inch\_030-away} & 00:07:45 & 1.1066 & 11.16 & 0.26 \\
    \code{BRJ\_gravel\_tracks\_on\_load\_full\_inch\_030-back} & 00:08:43 & 1.0852 & 11.06 & 0.26 \\
    \code{BRJ\_gravel\_tracks\_on\_load\_full\_inch\_060-away} & 00:03:50 & 1.2444 & 21.61 & 0.39 \\
    \code{BRJ\_gravel\_tracks\_on\_load\_full\_inch\_060-back} & 00:04:16 & 1.2224 & 20.50 & 0.39 \\
    \code{BRJ\_terrain\_tracks\_on\_load\_full\_inch\_030-test\_circuit} & 00:07:54 & 1.1031 & 42.38 & 0.66 \\
    \code{BRJ\_terrain\_tracks\_on\_load\_full\_inch\_040-test\_circuit} & 00:05:43 & 1.1642 & 59.22 & 1.04 \\
    \code{BRJ\_terrain\_tracks\_on\_load\_full\_inch\_020-test\_circuit} & 00:11:18 & 1.0392 & 19.67 & 0.31 \\
    \code{BRJ\_gravel\_tracks\_off\_load\_empty\_inch\_030-away} & 00:06:29 & 1.0938 & 6.36 & 0.18 \\
    \code{BRJ\_gravel\_tracks\_off\_load\_empty\_inch\_030-back} & 00:06:15 & 1.1005 & 6.91 & 0.19 \\
    \code{BRJ\_gravel\_tracks\_off\_load\_empty\_inch\_060-away} & 00:03:04 & 1.2392 & 15.73 & 0.35 \\
    \code{BRJ\_gravel\_tracks\_off\_load\_empty\_inch\_060-back} & 00:03:17 & 1.2252 & 16.49 & 0.37 \\
    \code{BRJ\_gravel\_tracks\_off\_load\_empty\_inch\_100-away} & 00:02:21 & 1.2954 & 21.07 & 0.37 \\
    \code{BRJ\_gravel\_tracks\_off\_load\_empty\_inch\_100-back} & 00:02:20 & 1.297 & 23.73 & 0.42 \\
    Stitched sequences of regular work (without pauses) & 11:20:00 & 0.43671 & 19.99 & 0.66 \\
    \hline
    \end{tabular}
    \label{tab:vibration_measures}
\end{table}

\begin{figure}
    \centering
    \includegraphics[width=0.9\linewidth]{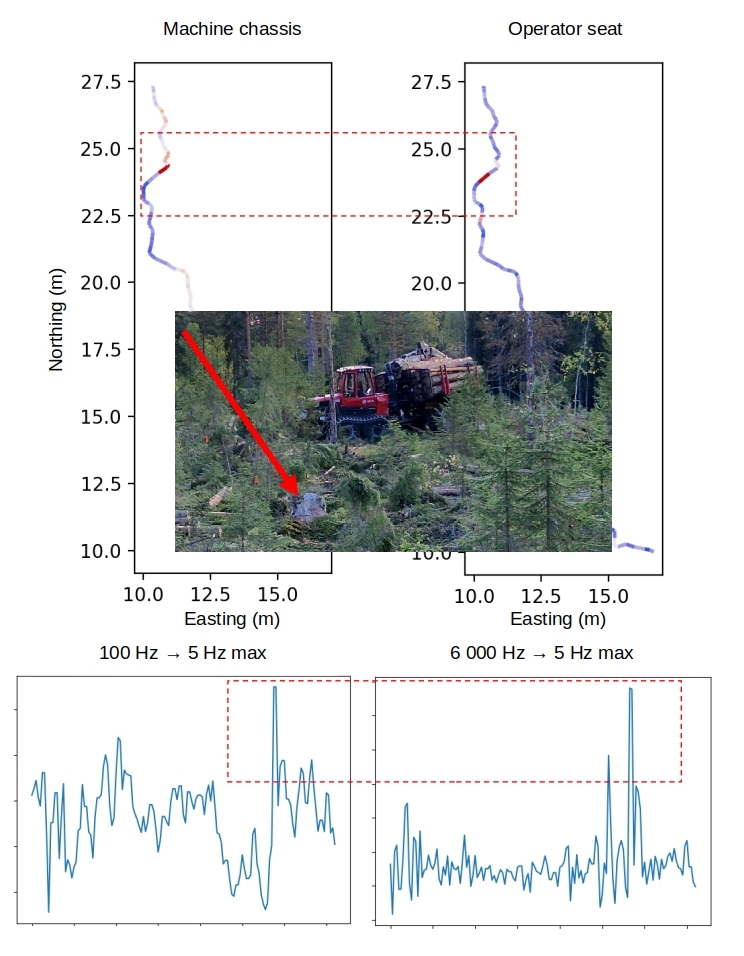}
    \caption{Downsampled machine vibration data (left) and operator seat vibration data (right) during the passage of a 90 cm high stone on the test circuit (between 20 and 25 m northing). Plotted along the machine path with high values in red and low values in blue and as a time series (bottom of the figure) with the stone passage indicated by a red dashed rectangle in both spatial and time series plots. Both datasets are accelerations in the vertical direction.}
    \label{fig:vibration}
\end{figure}

Using the maximum value for each data point as downsampling statistic, we plot the downsampled vibration data together with the machine accelerometer data for a subset of the test circuit covering the passage of a 90 cm high stone (Fig.~\ref{fig:vibration}).
In the vertical direction, peaks in acceleration are observed when the forwarder drives over obstacles, such as stones and stumps.
In this specific case, the vibration data shows a clear peak in the vertical direction when the forwarder drives down from the 90 cm high stone (Fig.~\ref{fig:vibration}).

\FloatBarrier

\subsection{StanForD production data}
\begin{figure}
    \centering
    \includegraphics[width=0.85\linewidth]{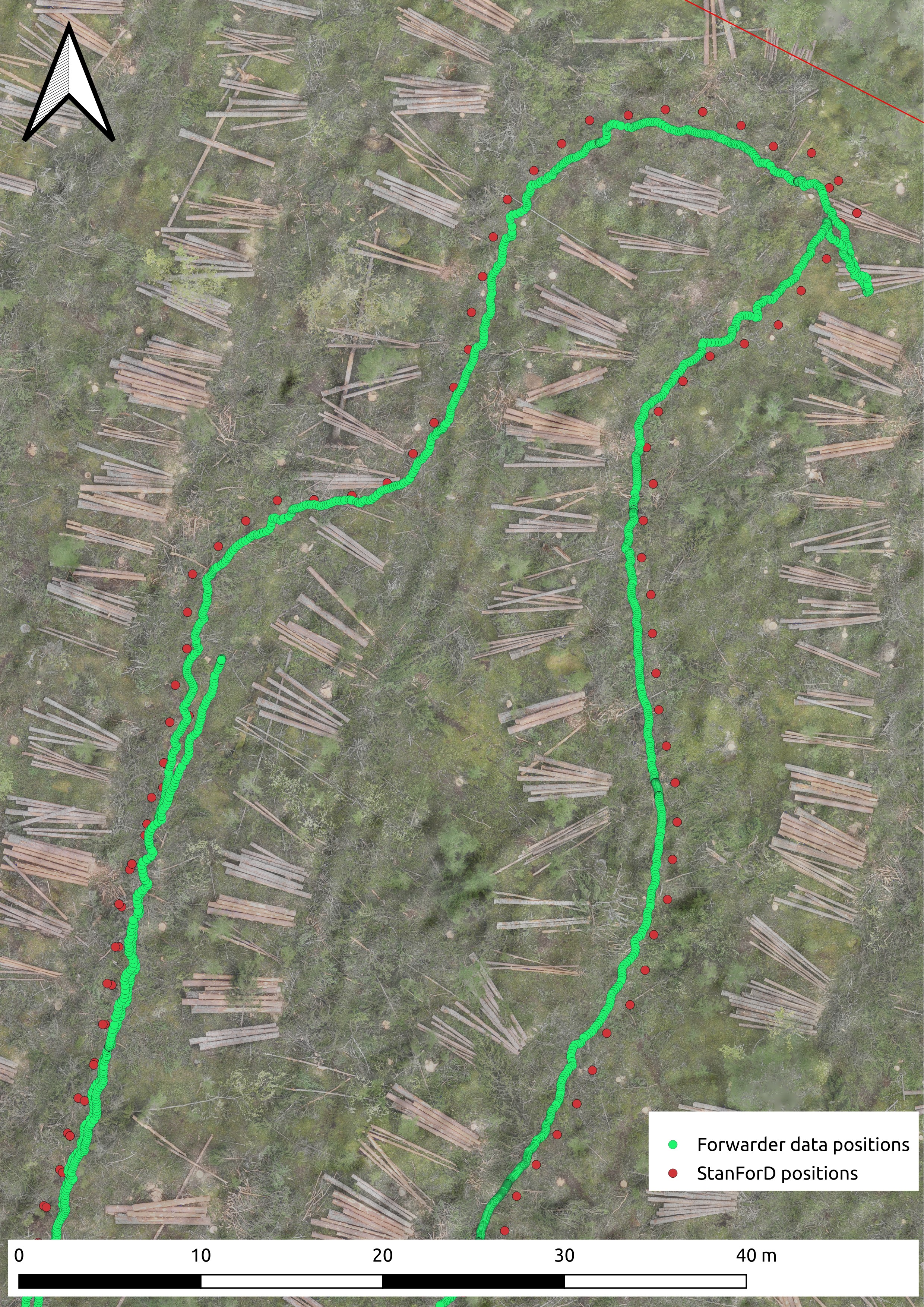}
    \caption{Machine positions from StanForD production data (red) and the RTK-GNSS log in the main dataset (green) plotted on a map of part of the Björsjö-site. The StanForD positions are upsampled to match the 5Hz time resolution of the other data. The background is the orthomosaic from the drone-footage made before forwarding.}
    \label{fig:stanford_map}
\end{figure}

The StanForD production data from the forwarder is available for the Björsjö-site and can be visualized and used in a variety of ways, for example machine positions in time series or on a map of the test site.
Since the machine positions are recorded at a coarser time-resolution in StanForD than the 5Hz used in the other data, the positions are upsampled to match the 5Hz time resolution by repeating identical positions until the next position is detected.
When plotting machine positions on a map, the difference in time resolution is noticeable (Fig.~\ref{fig:stanford_map}).

\begin{figure}
    \centering
    \includegraphics[height=0.40\textheight]{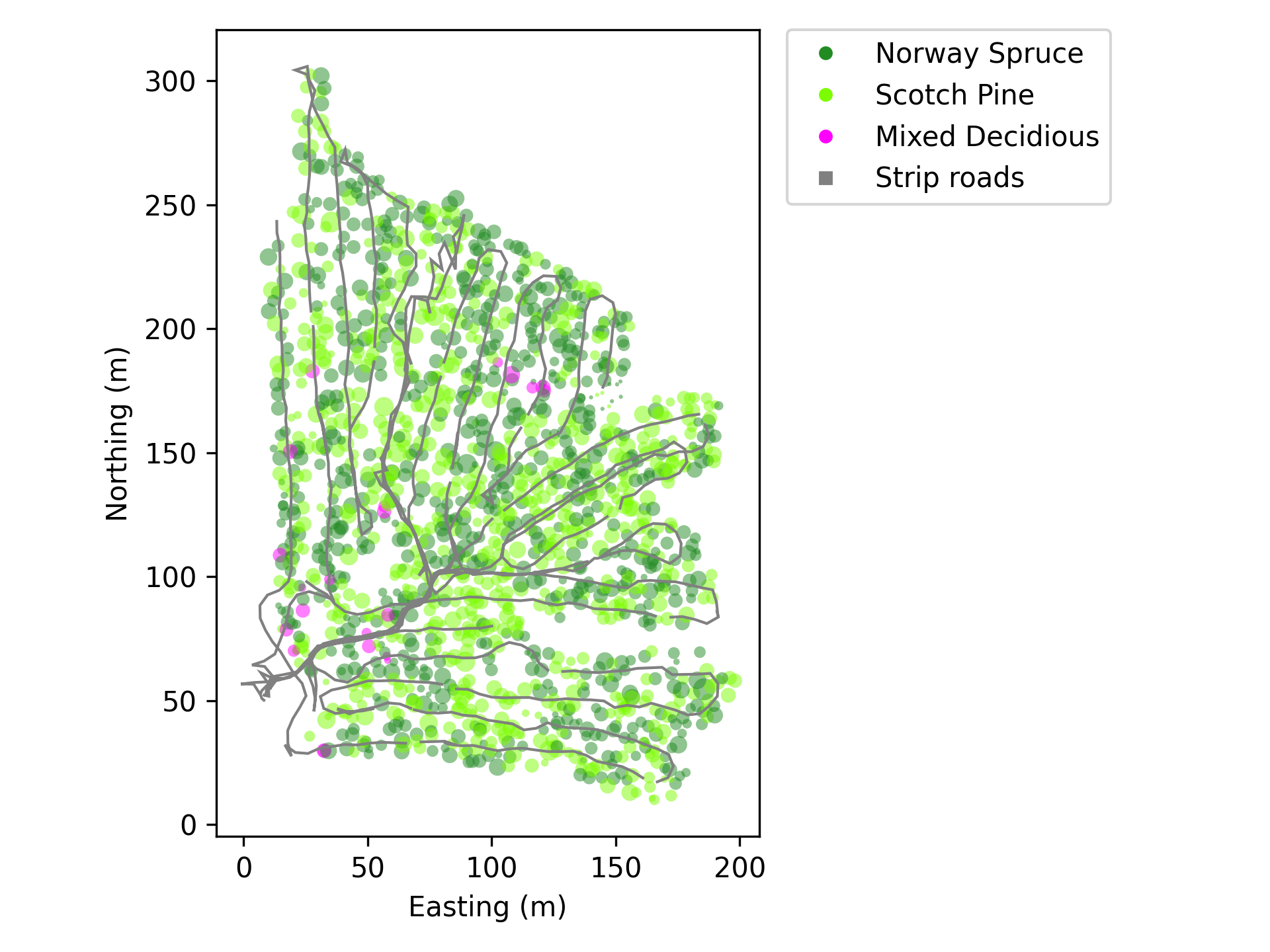}
    \caption{Stump positions of harvested trees in the forest at the Björsjö-site. The diameter of each tree in the plot corresponds to its diameter at breast height (DBH). Local northing and easting are shown in meters.}
    \label{fig:harvested_trees}
\end{figure}

\begin{figure}
    \centering
    \includegraphics[height=0.40\textheight]{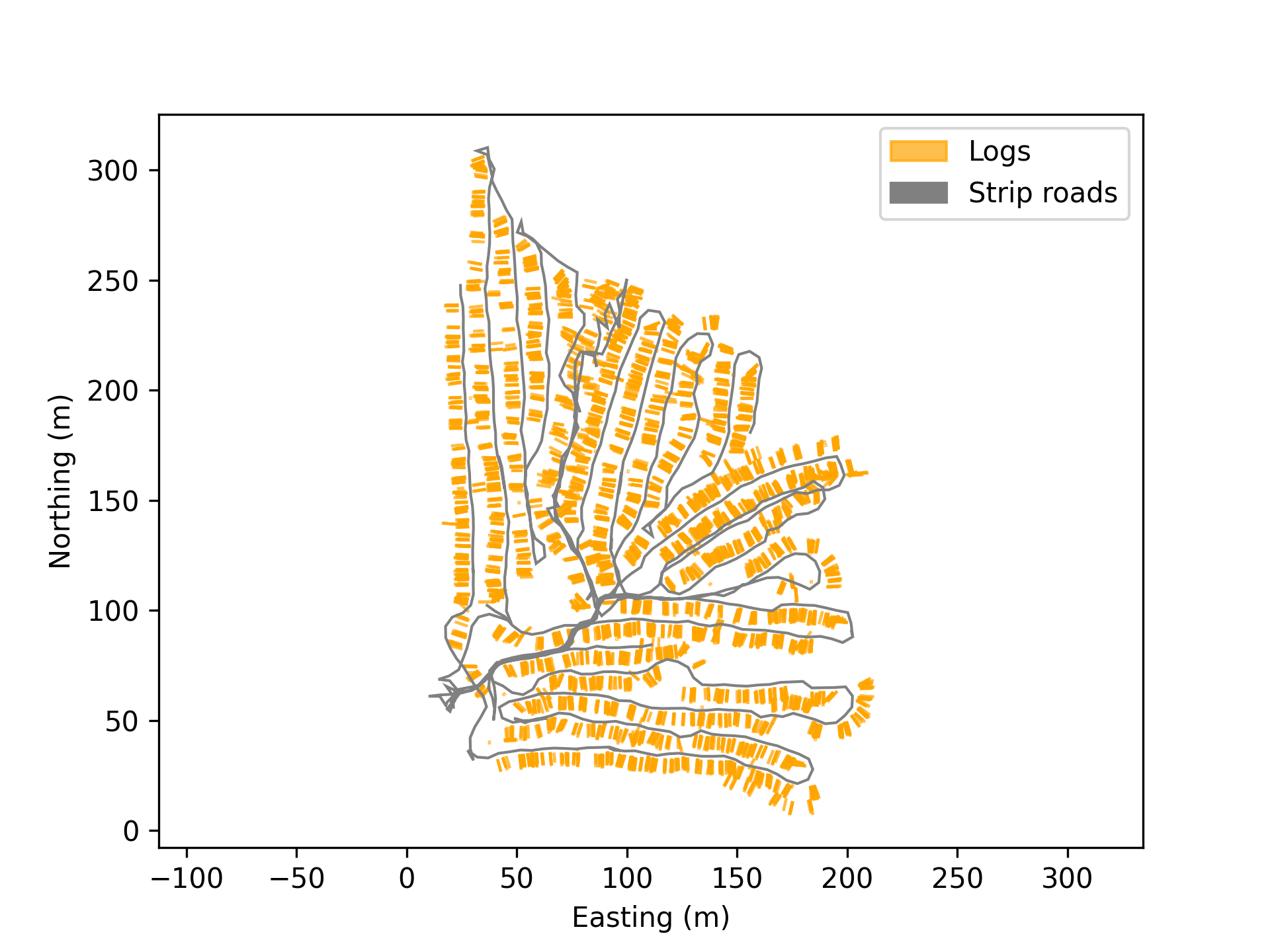}
    \caption{Positions of harvested logs in the forest at the Björsjö-site. Other information about each log could be added, such as if it is a sawlog or a pulplog. Local northing and easting are shown in meters.}
    \label{fig:harvested_logs}
\end{figure}

Production data from the harvester that was used to cut the trees before forwarding is available for both sites, Märrviken and Björsjö.
The harvester data are available in the same StanForD format as the forwarder data and can be visualized in the same way, however, the harvester data are of course not synchronized with the forwarder data.
Apart from machine positions, harvester data include other interesting information such as positions for harvested trees and processed logs, as well as tree species and dimensions.
The spatial distribution of logs is important for understanding the operative conditions for the forwarder.
The distribution of trees and logs can be visualized on a map of the test site (Fig.~\ref{fig:harvested_trees} and~\ref{fig:harvested_logs}).

\FloatBarrier

\section*{Limitations}
Some notes can be made about generalizability and expected accuracy for different parts of the dataset:
\begin{itemize}
    \item Since this dataset stems from experiments with one single machine operated by the two same persons throughout the experiments, generalizability may not be immediate, especially not to radically smaller forwarders. Our contribution is focused on providing as rich data as possible from these (time consuming) experiments, and hopefully more efforts can be made in the future to complete this kind of data to represent more machine sizes and brands.
    \item After further analysis of the driving speed it has become clear that there are more settings in the machine than Throttle position, Inch factor, and Gear that have effect on the target driving speed of the forwarder. There is a variable called transmission mode that controls the diesel engine rpm in different situations and thus the target driving speed, however this variable is not included in the dataset due to confidentiality reasons. No processing or analyses demonstrated in this article or the dataset repository rely on this (or any other) disclosed variable. It is only an example of data that might exist at the manufacturer but never were included in this dataset, data that might explain potential difficulties with specific further use of the dataset. The reason for the manufacturer not to release a full dataset for research with all available variables is that it would reveal too much to the competitors how the machines are engineered.
    \item GNSS-variables from the machine data, such as positions, heading, and speed are captured with a system developed by Komatsu Forest that uses two RTK-GNSS receivers on the cabin roof. When internet connection is available (most of the time) the system is connected to the Swedish SWEPOS network of base stations, enabling centimeter-level accuracy for positions. The dual receiver setup enables computation of heading. The accuracy variable value 'RTK\_Fix' in the dataset indicates optimal accuracy for the position and heading respectively.
    \item The video files are cut or muted to remove conversations held between tests.
    \item For pointcloud data, generated digital terrain models, and plotted machine paths, the coordinate reference system SWEREF99 - TM (EPSG:3006) is used. The experimental sites are situated relatively close to Sweden's central meridian, mitigating east-west related distortions.
    \item Drone orthomosaics from the Märrviken site have been manually corrected in horizontal directions by aligning visible objects with the terrain models produced from helicopter-borne LiDAR. This was necessary since the drone used in 2023 did not have any positional correction available and no manual control points were arranged in the field. The level of accuracy is sufficient for using the orthomosaic for visualization purposes together with the LiDAR data, for other uses it may be insufficient.
\end{itemize}

\section*{Ethics statement}
The authors confirm that they have read and followed the ethical requirements for publication in Data in Brief. The current work does not involve animal experiments or any data collected from social media platforms. Although machine operators and other people appear on video recordings they are not subjects of the study. Informed consent from all persons involved has been gathered and stored at Umeå University to comply with national legal demands of the open dataset. The consent includes being visible on video material. There are no limitations for reuse or redistribution of the dataset based on ethical considerations. Operator vibration measures were made on the operator seat, regardless of operator.

\section*{Credit author statement}
\textbf{Mikael Lundbäck:} Conceptualization, Methodology, Software, Validation, Formal analysis, Investigation, Data Curation, Writing - Original Draft, Writing - Review \& Editing, Visualization, Project administration.
\textbf{Erik Wallin:} Conceptualization, Methodology, Software, Validation, Formal analysis, Investigation, Data Curation, Writing - Original Draft, Writing - Review \& Editing, Visualization.
\textbf{Carola Häggström:} Conceptualization, Methodology, Validation, Investigation, Resources, Data Curation, Writing - Review \& Editing, Supervision.
\textbf{Mattias Nyström:} Conceptualization, Methodology, Validation, Investigation, Resources, Data Curation, Writing - Review \& Editing.
\textbf{Andreas Grönlund:} Methodology, Validation, Resources, Writing - Review \& Editing, Project administration.
\textbf{Mats Richardson:} Software, Validation, Formal analysis, Investigation, Data Curation, Writing - Review \& Editing.
\textbf{Petrus Jönsson:} Conceptualization, Methodology, Investigation, Resources, Writing - Review \& Editing, Project administration, Funding acquisition.
\textbf{William Arnvik:} Formal analysis, Investigation, Data Curation, Writing - Review \& Editing, Visualization.
\textbf{Lucas Hedström:} Software, Validation, Formal analysis, Resources, Data Curation, Writing - Original Draft, Writing - Review \& Editing, Visualization.
\textbf{Arvid Fälldin:} Conceptualization, Methodology, Software, Validation, Formal analysis, Investigation, Data Curation, Writing - Original Draft, Writing - Review \& Editing, Visualization, Project administration.
\textbf{Martin Servin:} Conceptualization, Methodology, Software, Validation, Formal analysis, Investigation, Resources, Writing - Original Draft, Writing - Review \& Editing, Visualization, Supervision, Project administration, Funding acquisition.

\section*{Acknowledgements}
This work was supported in part by Mistra Digital Forest (Grant DIA 2017/14 \#6) and Horizon Europe Project XSCAVE under Grant 101189836.
The authors would like to thank the forest company SCA for providing the aerial LiDAR data and machine- and personnel resources used in this study via collaborative field studies. We also thank Christian Höök, SLU, for his generous help with video data collection 2023.
Hans Pettersson, Umeå University, is acknowledged for his help with the vibration data collection and processing.

\section*{Declaration of competing interest}
Mattias Nyström and Andreas Grönlund are employed by Komatsu Forest AB, Umeå, Sweden. The authors declare that the research was conducted in the absence of any commercial, personal or financial relationships that could be construed as a potential conflict of interest.

\bibliographystyle{model1-num-names}

\end{document}